%% file: anonymous-submission-latex-2026.tex
\documentclass[letterpaper]{article} % DO NOT CHANGE THIS
\usepackage[]{aaai2026}  % DO NOT CHANGE THIS
\usepackage{times}  % DO NOT CHANGE THIS
\usepackage{helvet}  % DO NOT CHANGE THIS
\usepackage{courier}  % DO NOT CHANGE THIS
\usepackage[hyphens]{url}  % DO NOT CHANGE THIS
\usepackage{graphicx} % DO NOT CHANGE THIS
\usepackage{multirow}
\usepackage{natbib}  % DO NOT CHANGE THIS AND DO NOT ADD ANY OPTIONS TO IT
\usepackage{caption} % DO NOT CHANGE THIS AND DO NOT ADD ANY OPTIONS TO IT
\usepackage{algorithm}
\usepackage{algorithmic}
\usepackage{booktabs}
\usepackage{subcaption}
\usepackage{color}

\usepackage{newfloat}
\usepackage{listings}
\DeclareCaptionStyle{ruled}{labelfont=normalfont,labelsep=colon,strut=off} % DO NOT CHANGE THIS
\floatstyle{ruled}
\newfloat{listing}{tb}{lst}{}
\floatname{listing}{Listing}
\newcommand{\METHODNAME}{\textsc{MOSE}}

\title{Multiplicative Orthogonal Sequential Editing for Language Models}

\author{
    Hao-Xiang Xu\textsuperscript{\rm 1,2}\thanks{Equal contribution.}, Jun-Yu Ma\textsuperscript{\rm 1,2}\footnotemark[1], Ziqi Peng\textsuperscript{\rm 1}\footnotemark[1], Yuhao Sun\textsuperscript{\rm 1}, Zhen-Hua Ling\textsuperscript{\rm 1,2}, Jia-Chen Gu\textsuperscript{\rm 3}\thanks{Corresponding author.}
}
\affiliations{
    \textsuperscript{\rm 1} University of Science and Technology of China \\
    \textsuperscript{\rm 2}National Engineering Research Center of Speech and Language Information Processing \\
    \textsuperscript{\rm 3}University of California, Los Angeles

    \{nh2001620,mjy1999,aisis,syh3327\}@mail.ustc.edu.cn, zhling@ustc.edu.cn, gujc@ucla.edu
}

\usepackage{bibentry}
\usepackage{amsmath}  % 必须添加
\usepackage{cases}    % 如果需要更复杂的case环境
\usepackage{amsfonts}
\begin{document}

\maketitle

\begin{abstract}
Knowledge editing aims to efficiently modify the internal knowledge of large language models (LLMs) without compromising their other capabilities. 
The prevailing editing paradigm, which appends an update matrix to the original parameter matrix, has been shown by some studies to damage key numerical stability indicators (such as condition number and norm), thereby reducing editing performance and general abilities, especially in sequential editing scenario. 
Although subsequent methods have made some improvements, they remain within the additive framework and have not fundamentally addressed this limitation.
To solve this problem, we analyze it from both statistical and mathematical perspectives and conclude that multiplying the original matrix by an orthogonal matrix does not change the numerical stability of the matrix.
Inspired by this, different from the previous additive editing paradigm, a multiplicative editing paradigm termed \textbf{M}ultiplicative \textbf{O}rthogonal \textbf{S}equential \textbf{E}diting (\METHODNAME{}) is proposed. 
Specifically, we first derive the matrix update in the multiplicative form, the new knowledge is then incorporated into an orthogonal matrix, which is multiplied by the original parameter matrix. 
In this way, the numerical stability of the edited matrix is unchanged, thereby maintaining editing performance and general abilities.
We compared \METHODNAME{} with several current knowledge editing methods, systematically evaluating their impact on both editing performance and the general abilities across three different LLMs. Experimental results show that \METHODNAME{} effectively limits deviations in the edited parameter matrix and maintains its numerical stability. Compared to current methods, \METHODNAME{} achieves a 12.08\% improvement in sequential editing performance, while retaining 95.73\% of general abilities across downstream tasks. The code is available at {https://github.com/famoustourist/MOSE}.
\end{abstract}

\input{1_introduction}
\input{2_related}
\input{3_preliminary}
\input{4_analysis}

\input{5_method}
\input{6_experiments}
\input{7_conclusion}
\input{9_acknowledgements}

\bibliography{aaai2026}

\clearpage
\newpage
\appendix
\onecolumn

\input{8_appendix}

\end{document}

%% file: 1_introduction.tex
\section{Introduction}

While large language models (LLMs) have demonstrated impressive capabilities~\cite{DBLP:journals/corr/abs-2407-21783}, they frequently suffer from hallucinations caused by inaccurate or obsolete knowledge stored in their parameters~\cite{DBLP:journals/corr/abs-2309-01219}. Since retraining LLMs requires substantial time and resources, researchers have increasingly focused on \emph{knowledge editing} (a.k.a., \emph{model editing})~\cite{DBLP:conf/icml/MitchellLBMF22, DBLP:conf/nips/MengBAB22, DBLP:conf/iclr/MengSABB23, DBLP:conf/nips/0104L0XY0X0C24,DBLP:conf/icml/MaL0G24}.
Current approaches to knowledge editing can be roughly categorized into \textit{parameter-modifying} methods~\cite{DBLP:conf/nips/MengBAB22, DBLP:conf/iclr/MengSABB23} that directly modify a small subset of model parameters, or \textit{parameter-preserving} methods~\cite{DBLP:conf/emnlp/WangMDYSLGC024} that integrate additional modules without altering the model parameters. In this paper, we study the parameter-modifying editing methods.

\emph{Sequential knowledge editing}~\cite{DBLP:conf/emnlp/YaoWT0LDC023} allows LLMs to continually integrate new knowledge through consecutive updates.
In this scenario, existing editing methods predominantly follow an additive paradigm, directly modifying model parameters by adding update matrices.
However, previous studies~\cite{DBLP:conf/acl/HuC00024, DBLP:conf/iclr/MaWXLG25} have demonstrated that this additive editing paradigm leads the edited parameter matrix to deviate substantially from its original structure, severely damaging critical aspects of numerical stability, including the norm~\cite{kahan2013tutorial} and condition number~\cite{DBLP:journals/siammax/Sun00}.
These disruptions degrade the model's editing performance and general abilities.
Although methods like RECT~\cite{DBLP:conf/emnlp/GuXMLLCP24}, EAC~\cite{DBLP:conf/naacl/XuMLZG25}, PRUNE~\cite{DBLP:conf/iclr/MaWXLG25}, and AlphaEdit~\cite{DBLP:conf/iclr/FangJWMSW0C25} offer some mitigation, they only marginally extend the scope of editing and still cause damage to the numerical stability of the edited matrix, failing to fundamentally resolve the limitations inherent in the additive editing paradigm.
This impacts model scalability and presents major challenges for continuous learning in LLMs.

To address this challenge, we pose the following question: \textit{``Can we develop an editing paradigm that does not damage the numerical stability of the edited matrix, thereby maintaining both the editing performance and general abilities of the model?"}
Orthogonal transformations are numerically stable linear operations defined by orthogonal matrices. They encode rich semantic information by changing the angle of the vector~\cite{DBLP:conf/nips/LiuLLLYDS18}. 
In this paper, from a statistical perspective, compared to the previous editing paradigm based on additive updates, we first show that left-multiplying the matrix with an orthogonal matrix can strictly maintain its numerical stability indicator, including norm and condition number.
Furthermore, a rigorous mathematical analysis is provided demonstrating that such orthogonal transformations theoretically keep both the norm and the condition number of the matrix unchanged. 

\begin{figure*}
    \centering
\includegraphics[width=1\linewidth]{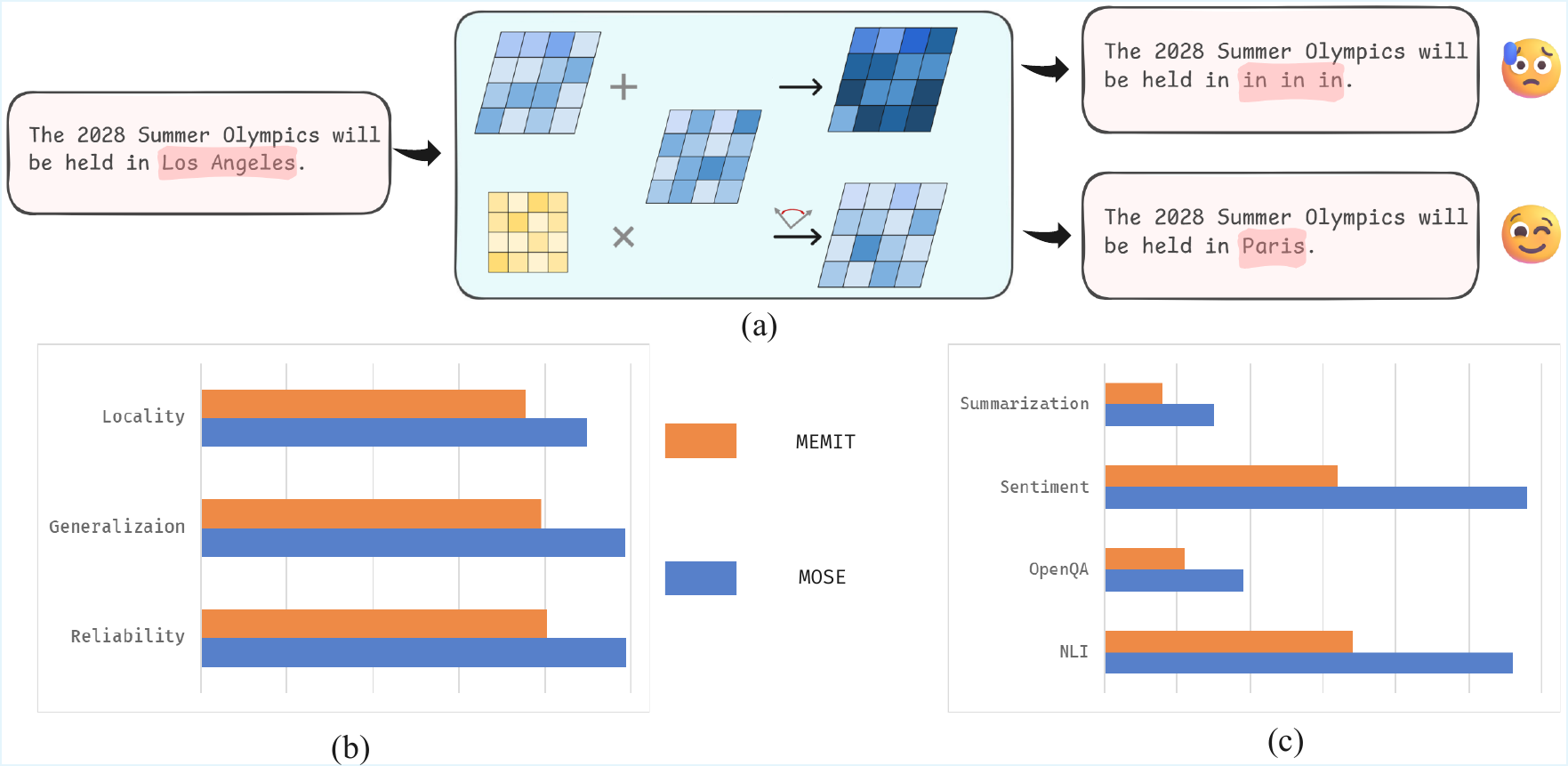}
    \caption{(a) Comparison between the previous methods and \METHODNAME{}. The previous methods performed updates by adding an update matrix, whereas the \METHODNAME{} employs left-multiplication by an orthogonal update matrix. (b) Comparison of editing performance after additive-based editing and after multiplicative-based editing by \METHODNAME{}. (c) Comparison of general downstream task performance before editing, after additive-based editing, and after multiplicative-based editing by \METHODNAME{}.}
    \label{head}
\end{figure*}
Inspired by these findings, we propose \textbf{M}ultiplicative \textbf{O}rthogonal \textbf{S}equential \textbf{E}diting (\METHODNAME{}), a multiplicative editing paradigm that leverages orthogonal transformations. 
Specifically, \METHODNAME{} introduces an orthogonal transformation update strategy, which injects new knowledge into an orthogonal update matrix and multiplies it with the original parameter matrix to perform knowledge editing, as shown in Figure~\ref{head}.
To achieve this, \METHODNAME{} explicitly minimizes the output error for both the target knowledge to be edited and the existing knowledge to be preserved.
By performing orthogonal transformation operations on the original matrix, the edited matrix not only stores the new knowledge but also does not affect its numerical stability.
This ensures that the knowledge within the model is not disturbed, thereby preserving the editing performance and the general abilities.

To validate the effectiveness of \METHODNAME{}, our study comprehensively evaluates the edited LLMs for both editing performance and general abilities in sequential editing. 
We evaluate \METHODNAME{} against \textbf{six popular knowledge editing methods}, including ROME~\cite{DBLP:conf/nips/MengBAB22}, MEMIT~\cite{DBLP:conf/iclr/MengSABB23}, RECT~\cite{DBLP:conf/emnlp/GuXMLLCP24}, EMMET~\cite{DBLP:conf/emnlp/GuptaBA24}, PRUNE~\cite{DBLP:conf/iclr/MaWXLG25} and AlphaEdit~\cite{DBLP:conf/iclr/FangJWMSW0C25}, across \textbf{three LLMs of varying sizes}, such as LLaMA3-8B~\cite{DBLP:journals/corr/abs-2407-21783}, LLaMA2-13B~\cite{DBLP:journals/corr/abs-2307-09288} and 
Qwen2.5-7B~\cite{DBLP:journals/corr/abs-2412-15115}. \textbf{Three editing datasets} across two types of editing data are selected to comprehensively evaluate the impact of knowledge editing on the editing performance of LLMs and \textbf{four representative tasks} are chosen to thoroughly assess how knowledge editing affects the general abilities of these models.
The experimental results demonstrate that during sequential editing, \METHODNAME{} surpasses existing approaches by 12.08\%, while still retaining 95.73\% of the general abilities.

In summary, our contributions to this paper are three-fold:
(1) This paper analyzes the key factor influencing both the editing performance and the general abilities of models after sequential editing, focusing on deviations in the parameter matrix and its numerical stability.
(2) A method termed \METHODNAME{} is proposed, which leverages orthogonal transformation to constrain deviations in the edited matrix and maintain its numerical stability.
(3) It is found that on models of different sizes, \METHODNAME{} demonstrates a 12.08\% improvement in editing performance compared to existing methods, while effectively preserving over 95.73\% of the model’s general abilities on downstream tasks.

%% file: 2_related.tex
\section{Related Work}

\paragraph{Knowledge Editing}
Current knowledge editing methods typically adopt either parameter modification or preservation strategies.  
\emph{Parameter-Modifying Methods} directly adjust model weights to inject new knowledge. Meta-learning approaches like KE~\cite{DBLP:conf/emnlp/CaoAT21}, MEND~\cite{DBLP:conf/iclr/MitchellLBFM22}, and InstructEdit~\cite{DBLP:conf/ijcai/0001T0LHXG0C24} use hypernetworks. Locate-then-edit methods, such as ROME~\cite{DBLP:conf/nips/MengBAB22}, compute updates via normal equations, while MEMIT~\cite{DBLP:conf/iclr/MengSABB23} scales this to batch editing.
\emph{Parameter-Preserving Methods} retain original weights through auxiliary designs. ICE~\cite{DBLP:conf/emnlp/ZhengLDFWXC23} uses in-context learning; SERAC~\cite{DBLP:conf/icml/MitchellLBMF22} employs external memory; T-Patcher~\cite{DBLP:conf/iclr/HuangSZZR023} and CaliNet~\cite{DBLP:conf/emnlp/DongDSXSL22} add editing-specific neurons. GRACE~\cite{DBLP:conf/nips/HartvigsenSPKG23} replaces hidden states with codebooks, while WISE~\cite{DBLP:conf/nips/0104L0XY0X0C24} integrates parameterized memory to enhance editing performance.

\paragraph{Sequential Editing}
Recent research systematically investigates the challenges in sequential model editing.  
\citet{DBLP:conf/emnlp/GuptaBA24} identifies ROME’s dual key vector design as a cause of editing failure, a finding extended by~\citet{DBLP:conf/emnlp/Yang0TMSYS24}, who attributes it to distributional discrepancies. \citet{DBLP:conf/acl/HuC00024} shows that representation overlap in whitening space hinders editing, while~\citet{DBLP:conf/iclr/MaWXLG25} theoretically links matrix condition numbers to editing instability. In response, recent methods propose targeted solutions: RECT~\cite{DBLP:conf/emnlp/GuXMLLCP24} applies sparse updates to limit parameter drift; PRUNE~\cite{DBLP:conf/iclr/MaWXLG25} constrains condition numbers; AlphaEdit~\cite{DBLP:conf/iclr/FangJWMSW0C25} leverages null-space constraints for near-lossless edits; and EAC~\cite{DBLP:conf/naacl/XuMLZG25} compresses editing anchors to preserve general abilities of the edited model.

Compared with previous studies~\cite{DBLP:conf/emnlp/GuXMLLCP24, DBLP:conf/iclr/MaWXLG25, DBLP:conf/naacl/XuMLZG25, DBLP:conf/iclr/FangJWMSW0C25} that are the most relevant to our work, a main difference should be highlighted. 
These approaches follow an additive paradigm, directly modifying model parameters by adding update matrices. However, this paradigm inevitably compromises the numerical stability of the edited matrix, ultimately degrading the model's editing performance and general abilities.
In contrast, our approach systematically leverages the numerical stability of orthogonal transformations by multiplying the original parameter matrix with an orthogonal update matrix. This operation preserves the edited matrix's numerical stability while maintaining strong editing performance and general abilities across multiple sequential edits.

%% file: 3_preliminary.tex
\section{Preliminary}
Knowledge editing enables targeted modification of knowledge encoded in language models (LMs) without full retraining, allowing adaptation to specific task requirements. This process facilitates precise updates to diverse learned representations, including but not limited to logical relationships, spatial understanding, and numerical reasoning.
In this work, we focus on structural knowledge editing represented as $(x_e, y_e)$\footnotemark{} pairs.
\footnotetext{Can be also represented as knowledge triple \( t = (subject, relation, object) \).}
Formally, given a language model $f_\theta \in \mathcal{F}$ that implements a mapping $f_\theta : \mathcal{X} \rightarrow \mathcal{Y}$ from inputs $x \in \mathcal{X}$ to predictions $y \in \mathcal{Y}$, editing aims to transform parameters $\theta \in \Theta$ to $\theta'$ such that $f_{\theta'}(x_e) = y_e$ when $f_\theta(x_e) \neq y_e$. The sequential editing extension involves an iterative process: for an edit set $\mathcal{E} = \left\{ (x_{ei}, y_{ei}) \mid i = 1, \ldots, n \right\}$ and initial model $f_{\theta_{0}}$, each step applies an editing function $K$ yielding $f_{\theta_i} = K(f_{\theta_{i-1}}, (x_{ei}, y_{ei}))$.

The effects of knowledge editing generally propagate through a neighborhood of inputs that are semantically connected to the modified knowledge, which is referred to \textit{editing scope}. To be considered successful, an editing operation must demonstrate two key properties: precise modification of outputs within this defined scope, and strict preservation of the model's behavior on all out-of-scope queries:
\[ 
f_{\theta_i}(x_{ei}) = 
\begin{cases} 
y_{ei} & \text{if} \,\, x_{ei} \in I(x_{ei}, y_{ei}), \\
f_{\theta_{i-1}}(x_{ei}) & \text{if} \,\, x_{ei} \in O(x_{ei}, y_{ei}). 
\end{cases} 
\]
The \textit{in-scope} \(I(x_{ei}, y_{ei})\) typically includes \(x_{ei}\) and its equivalence neighborhood \(N(x_{ei}, y_{ei})\), which encompasses related input/output pairs. In contrast, the \textit{out-of-scope} \(O(x_{ei}, y_{ei})\) comprises inputs unrelated to the edit example. 
Following established literature~\cite{DBLP:conf/emnlp/CaoAT21,DBLP:conf/iclr/MitchellLBFM22,DBLP:conf/nips/MengBAB22,DBLP:conf/iclr/MengSABB23,DBLP:conf/emnlp/YaoWT0LDC023}, we evaluate edits along three key dimensions: \emph{reliability}, \emph{generalization}, and \emph{locality}.

%% file: 4_analysis.tex
\begin{figure}[t]%[ht]
    \centering
        \includegraphics[width=1\linewidth]{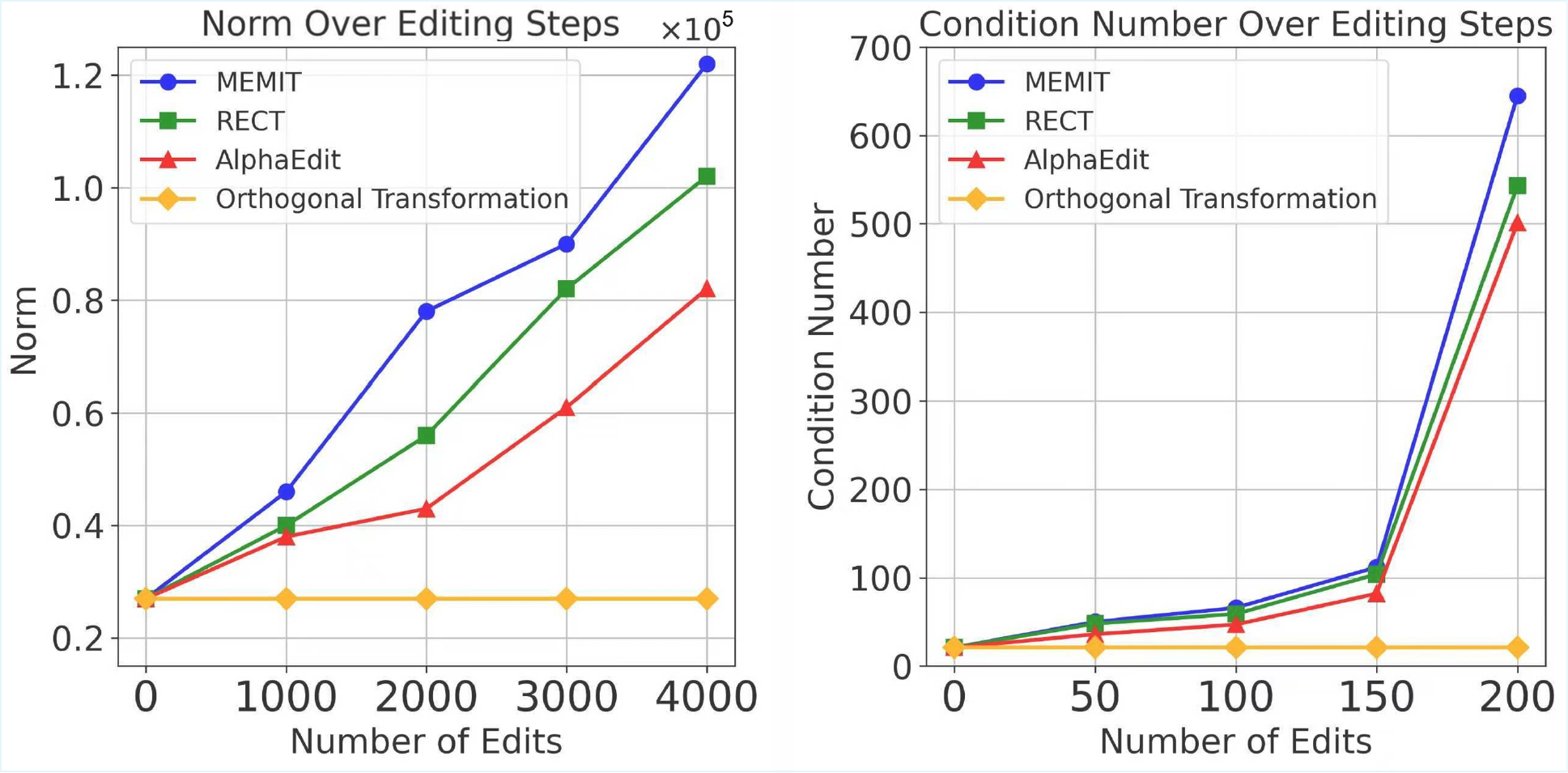}
    \caption{Illustration of the change of Frobenius norm and condition number in sequential editing at the edited layer using additive-based methods and orthogonal transformations. We selected LLaMA3-8B and CounterFact for experiments.}
    \label{analysis}
\end{figure}
\section{Analysis of Performance Degradation}
In this section, we statistically compare the additive editing methods with the orthogonal matrix left-multiplication approach. Additionally, we provide a rigorous mathematical analysis demonstrating that orthogonal transformations effectively preserve numerical stability by keeping both the norm and condition number of the matrix unchanged.

\subsection{Statistical Analysis}
Existing sequential editing methods based on additive update matrices typically take the following form:
\begin{equation}
W = W_0 + \Delta W_1 + \Delta W_2 + \cdots
\end{equation}
Previous studies have shown that existing methods undermine the numerical stability of the edited parameter matrix, causing significant increases in both the norm and condition number~\cite{DBLP:conf/iclr/MaWXLG25,DBLP:conf/naacl/XuMLZG25}. This destabilization negatively impacts the editing performance and general abilities of models.
The Frobenius norm of a matrix \( W \) can be defined as follows~\cite{kahan2013tutorial}:
\begin{equation}
\|W\|_F = \sqrt{ \sum_{i=1}^{m} \sum_{j=1}^{n} |W_{ij}|^2 },
\end{equation}
where \( W_{ij} \) denotes the element in the \( i \)-th row and \( j \)-th column of \( W \), and \( \|W\|_F \) represents the Frobenius norm, which measures the overall magnitude of the matrix by summing the squares of all its entries. 
The condition number \( \kappa_2(W) \) of a matrix \( W \) is~\cite{DBLP:journals/siammax/Sun00}:
\begin{equation}
\kappa_2(W) = \|W\|_2 \cdot \|W^{\dagger}\|_2,  
\end{equation}
where \( \|W\|_2 \) denotes the spectral norm of matrix \( W \), which is its largest singular value. \( W^{\dagger} \) represents the Moore-Penrose pseudoinverse of \( W \), and \( \|W^{\dagger}\|_2 \) equals the reciprocal of the smallest non-zero singular value of \( W \).

\paragraph{Additive Editing Methods} As shown in Figure~\ref{analysis}, editing methods such as ROME~\cite{DBLP:conf/nips/MengBAB22} and MEMIT~\cite{DBLP:conf/iclr/MengSABB23} cause a substantial increase in both the Frobenius norm and the condition number of the edited matrix, severely undermining its numerical stability. In contrast, methods like RECT~\cite{DBLP:conf/emnlp/GuXMLLCP24} and AlphaEdit~\cite{DBLP:conf/iclr/FangJWMSW0C25} demonstrate better capability in limiting edited matrix deviation and, to some extent, preserving its numerical stability. However, even these improved approaches still exhibit noticeable degradation when the number of edits becomes large. This degradation in numerical stability ultimately results in a pronounced decline in both the editing performance and the generalization capabilities of the model across downstream tasks.

\paragraph{Left-Multiplying Orthogonal Matrices} In contrast, we sequentially applied random orthogonal transformation by left-multiplying the edited matrix with a series of randomly generated orthogonal matrices at each editing step, thus simulating the sequential editing process:
\begin{equation}
    W_i = R_i W_{i-1} \quad \text{for } i = 1,\dots,k.
\end{equation}
As depicted in Figure~\ref{analysis}, this approach avoids significant growth in matrix norms and condition numbers while rigorously preserving the edited matrix's numerical stability.
Through this comparison, it becomes evident that additive update methods in existing knowledge editing frameworks tend to accumulate noise during sequential edits. While effective in early stages, they offer no mechanism to prevent the progressive distortion of the parameter matrix, leading to degradation in numerical properties such as the norm and condition number. In contrast, applying an orthogonal matrix through left multiplication helps preserve these properties, maintaining numerical stability even after a large number of edits, thereby enabling robust knowledge integration.

\subsection{Mathematical Analysis}
Furthermore, we provide a mathematical analysis showing that, due to the favorable properties of orthogonal transformations, our method is able to strictly preserve the numerical stability of the edited parameter matrix even after a large number of sequential edits. Specifically, orthogonal transformations are capable of keeping key numerical characteristics of matrices unchanged, including the Frobenius norm and the condition number. We begin by recalling the key properties of orthogonal matrices and then proceed with the formal proofs for each of the two quantities.
Let \( W \in \mathbb{R}^{m \times n} \) be a matrix, and let \( R \in \mathbb{R}^{m \times m} \) be an orthogonal matrix:
\begin{equation}
R^\top R = R R^\top = I,
\end{equation}
where \( I \) is the identity matrix. The orthogonal matrix \( R \) preserves vector lengths and angles.

\paragraph{Proof of Norm} 
Consider the matrix \( W' \) defined as \( W' = R W \). We want to prove that the Frobenius norm of \( W' \) is equal to that of \( W \):
\begin{equation}
\|W'\|_F = \sqrt{ \sum_{i=1}^{m} \sum_{j=1}^{n} |(RW)_{ij}|^2 }.
\end{equation}
Note that the Frobenius norm squared can be written as the trace of the matrix product:
\begin{equation}
\|W'\|_F^2 = \mathrm{tr} \left( (RW)^T (RW) \right).
\end{equation}
Since \( R \) is orthogonal, \( R^T R = I \), the identity matrix, thus:
\begin{equation}
\|W'\|_F^2 = \mathrm{tr} \left( W^T I W \right) = \mathrm{tr} \left( W^T W \right).
\end{equation}
Recognizing that \( \mathrm{tr}(W^T W) \) is exactly the squared Frobenius norm of \( W \), we obtain:
\begin{equation}
\|W'\|_F^2 = \|W\|_F^2.
\end{equation}
Taking the square root on both sides, we conclude:
\begin{equation}
\|W'\|_F = \|W\|_F.
\end{equation}
This shows that left-multiplying by an orthogonal matrix \( R \) does not change the Frobenius norm of \( W \).

\paragraph{Proof of Condition Number} Next, we prove that left-multiplying \( W \) by an orthogonal matrix does not change its condition number, even when \( W \) is not invertible.
Since \( R \) is orthogonal, the singular values of \( W' = R W \) are the same as those of \( W \), implying:
\begin{equation}
\|W'\|_2 = \|R W\|_2 = \|W\|_2.
\end{equation}
Moreover, by the properties of the Moore-Penrose pseudoinverse, we have:
\begin{equation}
W'^\dagger = (R W)^\dagger = W^\dagger R^\top.
\end{equation}
Since \( R^\top \) is orthogonal, it preserves the spectral norm, so:
\begin{equation}
\|W'^\dagger\|_2 = \|W^\dagger R^\top\|_2 = \|W^\dagger\|_2.
\end{equation}
Therefore, the condition number defined with the pseudoinverse satisfies:
\begin{equation}
\kappa_2(W') = \|W'\|_2 \cdot \|W'^\dagger\|_2 = \|W\|_2 \cdot \|W^\dagger\|_2 = \kappa_2(W).
\end{equation}
This shows that left-multiplying by an orthogonal matrix does not change the condition number of \( W \). Therefore, the above proof shows that orthogonal transformation preserve both the Frobenius norm and the condition number of a matrix, thereby strictly ensuring its numerical stability.

%% file: 5_method.tex
\section{Method}
Existing knowledge editing methods based on additive updates often undermine the numerical stability of the edited matrix, resulting in significant drops in both editing performance and general abilities of the model.
To address this issue, we propose a method named \textbf{M}ultiplicative \textbf{O}rthogonal \textbf{S}equential \textbf{E}diting (\METHODNAME{}), which left-multiplies the target parameter matrix with an orthogonal update matrix. \METHODNAME{} preserves editing performance and generalization by maintaining the numerical stability of the edited matrix, even after extensive sequential edits.

\subsection{Orthogonal Transformation-Based Updates}
Let \( W_0 \in \mathbb{R}^{d \times p} \) represent the original parameter matrix, where \( K_0 = \left[ k_1^0 \mid k_2^0 \mid \dots \mid k_{n_0}^0 \right] \in \mathbb{R}^{p \times n_0} \) is the matrix containing all the vectors whose representations we
want to preserve in a row, and \( K_E = \left[ k_1^e \mid k_2^e \mid \dots \mid k_E^e \right] \in \mathbb{R}^{p \times n_E} \) is the matrix containing a row of vectors representing the edits we are making in a batch. The target representation of \( K_E \) is denoted as \( V_E = \left[ v_1^e \mid v_2^e \mid \dots \mid v_E^e \right] \in \mathbb{R}^{d \times n_E} \).
We introduce an orthogonal transformation matrix \( R \in \mathbb{R}^{d \times d} \), which is constrained to satisfy:
\begin{equation}
R^\top R = R R^\top = I_d.
\end{equation}
This constraint ensures that \( R \) is an orthogonal matrix, preserving the geometric structure of the feature space.
Similar to MEMIT~\cite{DBLP:conf/iclr/MengSABB23}, our goal is to minimize the output error for both the knowledge being updated and the knowledge intended to be preserved:
\begin{equation}
 \min_{W} \lambda \| W K_0 - W_0 K_0 \|_F^2 + \| W K_E - V_E \|_F^2. 
\end{equation}
However, we seek to find an orthogonal matrix that operates directly on the original matrix, rather than computing an additive update matrix. Thus, the optimization problem is:
\begin{equation}
 \min_{R} \lambda \| R W_0 K_0 - W_0 K_0 \|_F^2 + \| R W_0 K_E - V_E \|_F^2. 
\label{optimize}
\end{equation}
Here, \( \lambda > 0 \) is a regularization parameter that controls the trade-off between preserving the original knowledge and memorizing the new knowledge. The first term ensures that the original knowledge representation remains unchanged, while the second term forces the transformed knowledge representation to match the desired output.

The optimization problem in Eq.~(\ref{optimize}) is a constrained least-squares problem.
When \( W_0 \) is fixed, the problem reduces to optimizing \( R \) for the best transformation:
\begin{equation}
\min_{R \in O(d)} \| R A - B \|_F^2,
\end{equation}
where \( A = \begin{bmatrix} \sqrt{\lambda} W_0 K_0 & W_0 K_E \end{bmatrix} \in \mathbb{R}^{d \times (n_0 + n_E)} \) and \( B = \begin{bmatrix} \sqrt{\lambda} W_0 K_0 & V_E \end{bmatrix} \in \mathbb{R}^{d \times (n_0 + n_E)} \).
This is a standard orthogonal Procrustes problem~\cite{schonemann1966generalized}, which has a closed-form solution. The optimal \( R \) is obtained via Singular Value Decomposition (SVD) of the matrix \( M = B A^\top \):
\begin{equation}
M = U \Sigma V^\top. 
\end{equation}
Thus, the optimal solution for \( R \) is:
\begin{equation}
R = U V^\top.
\end{equation}
By multiplying the obtained orthogonal matrix \( R \) with the original parameter matrix \( W_0 \), we have successfully completed the knowledge update. Specifically, we employ the method proposed in previous work to derive the components \( K_0 \), \( K_E \), and \( V_E \)~\cite{DBLP:conf/nips/MengBAB22, DBLP:conf/iclr/MengSABB23}, which play crucial roles in the optimization process. For detailed definitions and calculations of these components, the reader is referred to the Appendix.

\renewcommand{\arraystretch}{0.6}
\begin{table*}
    \centering
    
    \begin{tabular}{lccccccc}
        \toprule
        \multirow{2}{*}{Method} & \multirow{2}{*}{Model} & \multicolumn{3}{c}{\textbf{CounterFact}} & \multicolumn{3}{c}{\textbf{ConceptEdit-Inter}}\\
        \cmidrule(lr){3-5} \cmidrule(lr){6-8}
        & & Reliability & Generalization & Locality & Reliability & Generalization & Locality \\
        \midrule
        % ROME & \multirow{6}{*}{LLama3-8B} & 0.0000 & 0.0000 & 0.0000 & 0.0000 & 0.0000 & 0.0000 \\
        ROME & \multirow{6}{*}{\rotatebox{90}{LLama3-8B}} & 0.0000 & 0.0000 & 0.0000 & 0.0000 & 0.0000 & 0.0000 \\
        MEMIT & & 0.0000 & 0.0000 & 0.0000 & 0.0000 & 0.0000 & 0.0000 \\
        RECT & & 0.5688 & 0.3288 & 0.2517 & 0.3531 & 0.2084 & 0.1502 \\
        EMMET & & 0.6398 & 0.4937 & 0.3253 & 0.3877 & 0.2258 & 0.1724 \\
        PRUNE & & 0.7886 & 0.7140 & 0.5763 & 0.6295 & 0.4414 & 0.3193 \\
        AlphaEdit & & 0.9018 & 0.8260 & 0.7831 & 0.7012 & 0.6007 & 0.5212 \\
        \midrule
        \METHODNAME{} & & \textbf{0.9887} & \textbf{0.9863} & \textbf{0.8972} & \textbf{0.7859} & \textbf{0.7275} & \textbf{0.6856} \\
        \midrule
        \midrule
        % ROME & \multirow{6}{*}{Qwen2.5-7B} & 0.0000 & 0.0000 & 0.0000 & 0.0000 & 0.0000 & 0.0000 \\
        ROME & \multirow{6}{*}{\rotatebox{90}{Qwen2.5-7B}} & 0.0000 & 0.0000 & 0.0000 & 0.0000 & 0.0000 & 0.0000 \\
        MEMIT & & 0.0000 & 0.0000 & 0.0000 & 0.0000 & 0.0000 & 0.0000 \\
        RECT & & 0.6203 & 0.4745 & 0.3582 & 0.3737 & 0.2306 & 0.1738 \\
        EMMET & & 0.6702 & 0.5589 & 0.4771 & 0.4593 & 0.2641 & 0.1903 \\
        PRUNE & & 0.8115 & 0.7860 & 0.6823 & 0.6708 & 0.5009 & 0.4120 \\
        AlphaEdit & & 0.9519 & 0.9241 & 0.8418 & 0.7346 & 0.6453 & 0.6116 \\
        \midrule
        \METHODNAME{} & & \textbf{0.9981} & \textbf{0.9902} & \textbf{0.9098} & \textbf{0.8012} & \textbf{0.7547} & \textbf{0.7069} \\
        \midrule
        \bottomrule
    \end{tabular}
    \caption{In the single-sequential editing scenario, the editing performance of different methods on CounterFact and ConceptEdit-Inter. We performed 4000 sequential edits.}
    \label{single-sequential-t}
\end{table*}

\subsection{Layer Selection Algorithm}
Inspired by previous work~\cite{DBLP:conf/acl/HuC00024, DBLP:conf/nips/QiuLFXFLZWS23}, we believe that different knowledge resides in different layers of the model, and that selectively modifying specific layers for specific knowledge can enhance editing performance.
When editing new knowledge, we first analyze which layer exhibits the strongest activation in response to the specific knowledge. This activation strength reflects how much the output of each layer responds to the given input and is mathematically formulated as:
\begin{equation}
\arg\max_l \left\| \sigma\left(\mathbf{x} \cdot W_{fc}^l\right) \right\|_2,    
\end{equation}
where \( W_{fc}^l \) denotes the weight matrix of the feed-forward network (FFN) in the \( l \)-th layer of the transformer, and \(\sigma(\cdot)\) is the non-linear activation function applied element-wise. This operation captures the transformed representation of the input \(\mathbf{x}\) and helps identify the layer most relevant to the knowledge being edited.
Considering both how strongly a layer reacts to the target knowledge, we define the final editing layer as follows:
\begin{equation}
l^* = \arg\min_l \left\|
\frac{V_E^l - W_0^l K_E^l}{\left\| W_0^l \right\|_2 \cdot \sigma(\mathbf{x} \cdot W_{fc}^l)}
\right\|_F,
\end{equation}
where the denominator \(\| W_{0}^l \|_2\) acts as a normalization term, facilitating fair comparison across layers. Ultimately, we identify the editing target as \( W_{0}^{l^*} \) at layer \( l^* \).
In addition, we found that editing multiple layers simultaneously can further enhance editing performance. Therefore, once we identify the target layer for editing, we modify that layer along with its two adjacent layers. In later sections, we present corresponding ablation experiments and provide further analysis to support this proposed approach.

%% file: 6_experiments.tex
\begin{figure*}[ht]
    \centering
    \begin{subfigure}{0.8\linewidth}
        \centering
        \includegraphics[width=\linewidth]{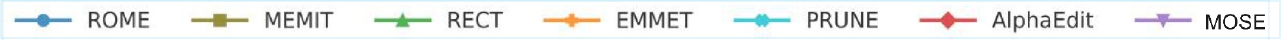}
    \end{subfigure}
    \hfill
    \begin{subfigure}{1\linewidth}
        \centering
        \includegraphics[width=\linewidth]{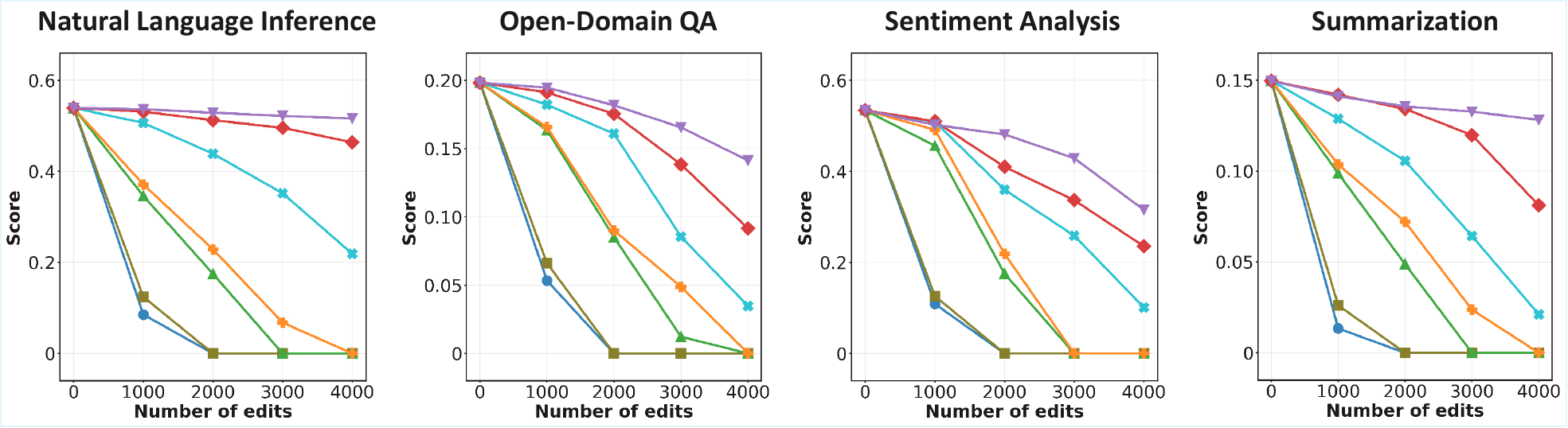}
    \end{subfigure}
    \caption{Edited on the CounterFact dataset, the general task performance of varying methods with LLaMA3-8B as the number of edits increases, under the single-sequential scenario.}
    \label{single-sequential-f}
\end{figure*}

\section{Experiments}

\subsection{Experimental Setup}
Experiments were conducted on three LLMs: LLaMA3-8B~\cite{DBLP:journals/corr/abs-2407-21783}, LLaMA2-13B~\cite{DBLP:journals/corr/abs-2307-09288}, and Qwen2.5-7B~\cite{DBLP:journals/corr/abs-2412-15115}. The baseline editing methods used were ROME~\cite{DBLP:conf/nips/MengBAB22}, MEMIT~\cite{DBLP:conf/iclr/MengSABB23}, RECT~\cite{DBLP:conf/emnlp/GuXMLLCP24}, EMMET~\cite{DBLP:conf/emnlp/GuptaBA24}, PRUNE~\cite{DBLP:conf/iclr/MaWXLG25} and AlphaEdit~\cite{DBLP:conf/iclr/FangJWMSW0C25}. The editing performance was evaluated using two types of datasets: factual knowledge-based datasets, including ZsRE~\cite{DBLP:conf/conll/LevySCZ17} and CounterFact~\cite{DBLP:conf/nips/MengBAB22}, and a conceptual knowledge-based dataset, ConceptEdit~\cite{DBLP:conf/emnlp/WangMDYSLGC024}. The models were assessed based on metrics such as reliability, generalization, and locality~\cite{DBLP:conf/nips/MengBAB22, DBLP:conf/iclr/MengSABB23, DBLP:conf/emnlp/YaoWT0LDC023, DBLP:conf/naacl/XuMLZG25}. To measure the general abilities of the models before and after editing, four downstream tasks were selected: \textbf{natural language inference}~\cite{DBLP:conf/mlcw/DaganGM05}, \textbf{summarization}~\cite{gliwa-etal-2019-samsum}, \textbf{open-domain question-answering}~\cite{DBLP:journals/tacl/KwiatkowskiPRCP19}, and \textbf{sentiment analysis}~\cite{DBLP:conf/emnlp/SocherPWCMNP13}. Due to page limitations, the experiments presented in our main text include only a representative subset of the results. Readers can refer to the Appendix for further details on the experimental setups and more comprehensive results.
\renewcommand{\arraystretch}{0.6}
\begin{table*}
    \centering
\begin{tabular}{lccccccc}
        \toprule
        \multirow{2}{*}{Method} & \multirow{2}{*}{Model} & \multicolumn{3}{c}{\textbf{CounterFact}} & \multicolumn{3}{c}{\textbf{ConceptEdit-Inter}}\\
        \cmidrule(lr){3-5} \cmidrule(lr){6-8}
        & & Reliability & Generalization & Locality & Reliability & Generalization & Locality \\
        \midrule
        % ROME & \multirow{7}{*}{LLama3-8B}  & 0.0000 & 0.0000 & 0.0000 & 0.0000 & 0.0000 & 0.0000 \\
        ROME & \multirow{7}{*}{\rotatebox{90}{LLama3-8B}}  & 0.0000 & 0.0000 & 0.0000 & 0.0000 & 0.0000 & 0.0000 \\
        MEMIT &  & 0.0000 & 0.0000 & 0.0000 & 0.0000 & 0.0000 & 0.0000 \\
        RECT &  & 0.5266 & 0.3075 & 0.2382 & 0.3234 & 0.1993 & 0.1397 \\
        EMMET &  & 0.6287 & 0.4695 & 0.3114 & 0.3866 & 0.2178 & 0.1563 \\
        PRUNE &  & 0.7738 & 0.6899 & 0.5190 & 0.5682 & 0.4097 & 0.3083 \\
        AlphaEdit &  & 0.8222 & 0.7835 & 0.7091 & 0.6981 & 0.5928 & 0.4977 \\
        \midrule
        \METHODNAME{} &  & \textbf{0.9422} & \textbf{0.9361} & \textbf{0.8819} & \textbf{0.7585} & \textbf{0.7191} & \textbf{0.6475} \\
        \midrule
        \midrule
        % ROME & \multirow{7}{*}{Qwen2.5-7B}  & 0.0000 & 0.0000 & 0.0000 & 0.0000 & 0.0000 & 0.0000 \\
        ROME & \multirow{7}{*}{\rotatebox{90}{Qwen2.5-7B}}  & 0.0000 & 0.0000 & 0.0000 & 0.0000 & 0.0000 & 0.0000 \\
        MEMIT &  & 0.0000 & 0.0000 & 0.0000 & 0.0000 & 0.0000 & 0.0000 \\
        RECT &  & 0.6080 & 0.4679 & 0.3239 & 0.3659 & 0.2266 & 0.1593 \\
        EMMET &  & 0.6579 & 0.5125 & 0.4338 & 0.4402 & 0.2461 & 0.1751 \\
        PRUNE &  & 0.7707 & 0.7527 & 0.6300 & 0.6395 & 0.4800 & 0.3857 \\
        AlphaEdit &  & 0.9427 & 0.8911 & 0.8094 & 0.6733 & 0.6162 & 0.5743 \\
        \midrule
        \METHODNAME{} &  & \textbf{0.9775} & \textbf{0.9514} & \textbf{0.9031} & \textbf{0.7526} & \textbf{0.7258} & \textbf{0.6731} \\
        \midrule
        \bottomrule
    \end{tabular}
    \caption{In the batch-sequential editing scenario, the editing performance of different methods on CounterFact and ConceptEdit-Inter. We set the batch size to 10 and performed 500 sequential edits.}
    \label{batch-sequential-t}
\end{table*}

\subsection{Results in Single-Sequential Editing Scenarios}
This section illustrates, in the single-sequential editing scenario, the editing performance and general abilities across downstream tasks of the edited model.

\paragraph{Editing Performance}
We aim for the sequentially edited model to retain all previously edited knowledge. To evaluate this, we performed a sequence of edits on the model and assessed its editing performance on the edited knowledge, which serves as an indicator of the model’s retention ability. Table~\ref{single-sequential-t} reports the editing performance of LLaMA3-8B and Qwen2.5-7B after 4000 sequential edits using both the CounterFact and Concept-Inter datasets across different editing methods. We observe that under previous editing methods, the model's performance deteriorates significantly as the number of sequential edits increases, regardless of the type of knowledge being edited. This suggests that these methods may introduce substantial disruptions to the model, reducing its ability to preserve previously integrated knowledge. In contrast, \METHODNAME{} demonstrates stronger retention performance in single-sequence editing scenarios by effectively maintaining the numerical stability of the model’s parameter matrix throughout the editing process.

\paragraph{General Abilities}
Applying CounterFact as the editing dataset, Figure~\ref{single-sequential-f} shows the task performance of different editing methods on LLaMA3-8B. It can be observed that when sequential edits are performed using traditional methods, the general abilities of the edited model fluctuate significantly and tend to decline as the number of edits increases. This issue becomes especially pronounced when the number of edits is large, severely limiting the model’s scalability. In contrast, by minimizing deviations between the pre- and post-edited models, \METHODNAME{} effectively preserves the general abilities of the edited model across downstream tasks.

\subsection{Results in Batch-Sequential Editing Scenarios}
To further demonstrate the superiority and robustness of our method, we introduce a more challenging scenario—\textit{batch-sequential editing}—which requires modifying multiple pieces of knowledge during each editing step. This section comprehensively illustrates the editing performance of the edited models. Using \textsc{CounterFact} and \textsc{ConceptEdit-Inter} as the editing datasets, Table~\ref{batch-sequential-t} presents a comprehensive comparative evaluation of the editing performance of \textsc{LLaMA3-8B} and \textsc{Qwen2.5-7B} using different methods under the batch-sequential setting. In these experiments, we set the batch size to 10 and performed 500 editing steps, evaluating each model’s ability to retain all previously edited knowledge. The results show that, under prior editing methods, simultaneously editing multiple knowledge pieces at each step causes more severe degradation than editing one piece at a time, leading to diminished overall performance and revealing that previous knowledge-editing methods face greater challenges in this scenario. In contrast, \METHODNAME{} consistently maintains strong editing performance under batch-sequential editing and demonstrates superior scalability compared with prior approaches. More experiments of general abilities can refer to the Appendix.

\begin{figure}[ht]
    \centering
    \includegraphics[width=0.85\linewidth]{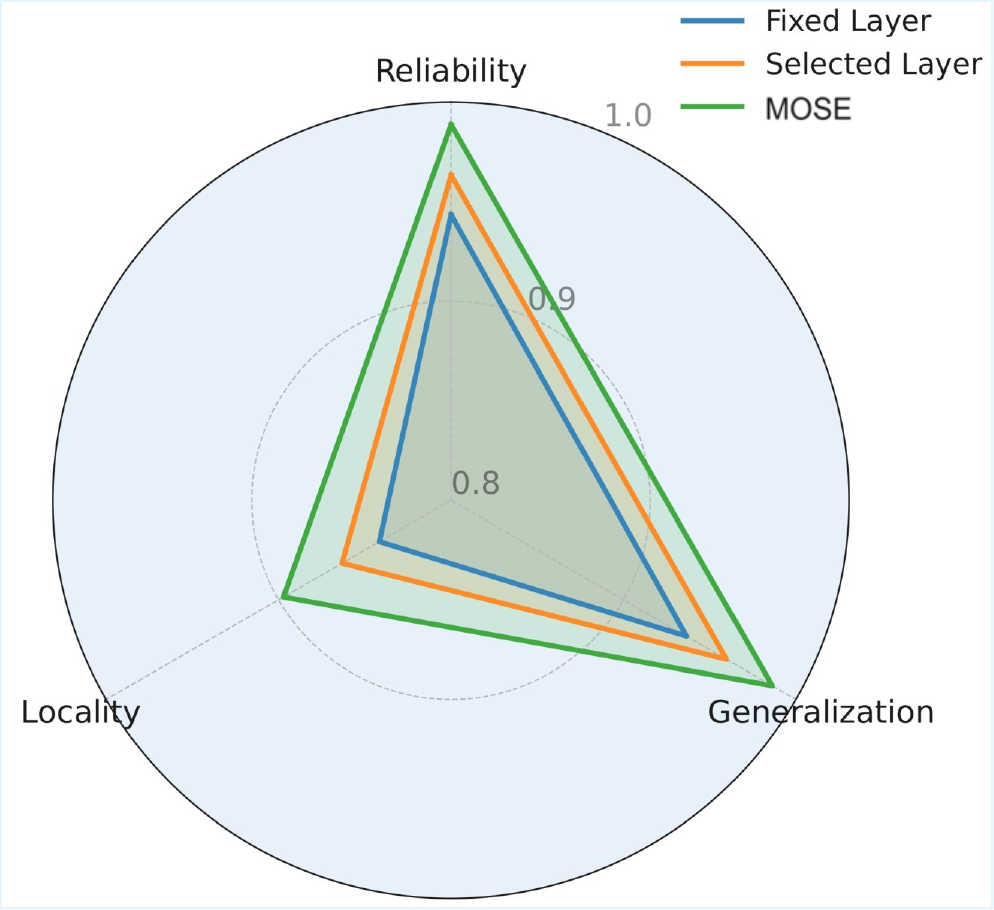}
    \caption{Ablation analysis of editing performance in the batch-sequential scenario. The experiment was conducted with LLaMA3-8B on CounterFact dataset.}
    \label{aba}
    \vspace{-3mm}
\end{figure}

\subsection{Ablation Study}
To evaluate the effectiveness of the layer selection algorithm in \METHODNAME{}, we conducted a series of ablation studies under three experimental setups. In the first setup, editing is performed on a predefined layer, where the number of layer matches it used in the ROME baseline. In the second setup, editing is applied to a single layer that is selected using our layer selection algorithm. In the third setup, which corresponds to the complete \METHODNAME{}, editing is performed on the selected layer as well as its immediate neighboring layers.
As shown in Figure~\ref{aba}, editing a predefined layer yields the weakest performance, indicating that knowledge is distributed across layers rather than confined to a single one. Thus, fixed-layer strategies fall short in handling diverse knowledge. Selecting a single layer via a layer selection algorithm improves results, but \METHODNAME{}, which edits both the selected and neighboring layers, performs best in batch-sequential settings.

\begin{figure}[ht]
    \centering
    \includegraphics[width=1\linewidth]{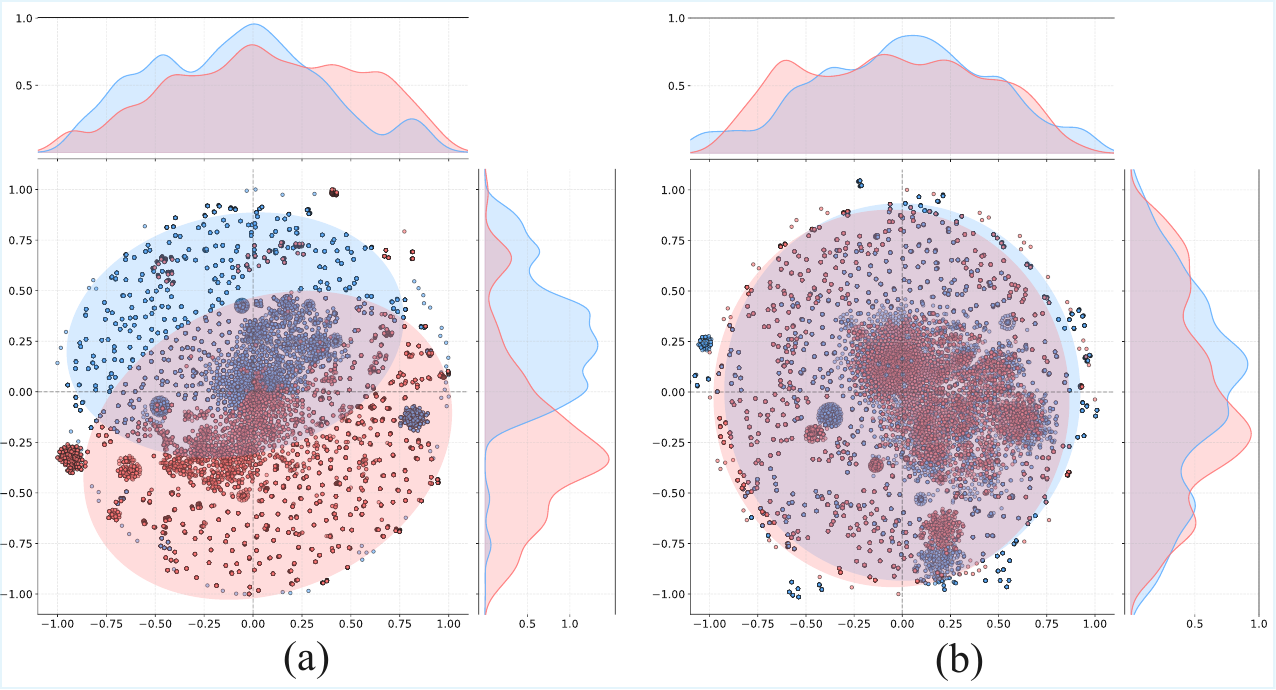}
    \caption{The t-SNE visualization of the hidden states of the LLaMA3-8B before and after editing, where (a) corresponds to edits made using RECT, and (b) corresponds to edits made using \METHODNAME{}. The top and right curves show the marginal distributions of the two t-SNE results.}
    \label{visual-SNE}
    \vspace{-3mm}
\end{figure}

\subsection{Further Analysis}
Building on the empirical results in the previous sections, we further investigate why \METHODNAME{} outperforms prior editing methods in maintaining both editing performance and general abilities. We begin with a guiding question: \emph{Does injecting a new fact via model editing alter the model’s downstream behavior?} Ideally, outputs for content unrelated to the edit should remain indistinguishable from the pre-edit state. When an edit is incorporated into a target layer through newly formed key--value pairs \( (k_\ast, v_\ast) \), the signal carried by \( v_\ast \) can propagate through attention and decoding, potentially reshaping later hidden states and, ultimately, the prediction. This perspective suggests a simple diagnostic: compare the representation geometry for the same inputs before and after stacking many edits.
To examine this, we design two complementary experiments. From a set of \(2000\) edits, we cache the per-edit parameter increments \( \Delta W_i \). We then evaluate under two matched conditions using the same inputs: (i) applying only the single update \( \Delta W_i \) and recording the final-layer representations of relation tokens for the corresponding fact, and (ii) applying the sequentially accumulated update \( \Delta W_{\text{total}}=\sum_i \Delta W_i \) and collecting the same representations again. We visualize both settings with t\textsc{-}SNE~\citep{DBLP:conf/emnlp/SocherPWCMNP13} and to guard against projection artifacts.

As shown in Fig~\ref{visual-SNE}(a), when the number of edits is large, RECT yields clearly separated distributions between the “single-update” and “cumulative-update” regimes. This separation indicates that stacking many edits reshapes the model’s outputs and thereby erodes both editing performance and general abilities on ostensibly unrelated content. In contrast, as shown in Fig~\ref{visual-SNE}(b), \METHODNAME{} produces two distributions that nearly overlap, with closely matched marginal statistics. This behavior suggests that even after extensive sequential editing, the output distribution for each edited fact remains stable and localized to the intended directions, rather than spilling over into unrelated representations.

\begin{table}[t]
    \centering
    \caption{Comparison of average editing time per sample (seconds) with different methods}
    \resizebox{0.45\textwidth}{!}{
        \begin{tabular}{c|ccccc}
        \toprule
        \textbf{Method} & ROME & MEMIT & RECT & AlphaEdit & \METHODNAME{} \\
        \midrule
        \textbf{Time(s)} & 12.09 & 16.16 & 12.88 & 18.53 & 18.60 \\
        \bottomrule
        \end{tabular}
    }
    \label{tab:placeholder}
\end{table}

\subsection{Time Costs}
We evaluated the runtime of various methods on the CounterFact, with results summarized in Table~\ref{tab:placeholder}. All experiments were conducted on a single A800-80G GPU. Compared with other additive-based editing methods, our approach requires slightly more editing time, mainly due to the computation of Singular Value Decomposition (SVD). However, given the significant improvements in both stability and overall performance, we consider this time–performance trade-off to be reasonable and worthwhile.

%% file: 7_conclusion.tex
\section{Conclusion}
This paper addresses sequential model editing and shows that existing additive update methods cause deviations and numerical instability in the parameter matrix. Through statistical and mathematical analysis, we demonstrate that orthogonal transformations preserve numerical properties during editing. Based on this, we propose \METHODNAME{}, which maintains stability by multiplicatively applying orthogonal transformations to the original matrix. Experiments confirm that \METHODNAME{} effectively maintains numerical stability, thereby preserving both editing performance and general abilities. 

%% file: 9_acknowledgements.tex
\section*{Acknowledgements}

This work is partially funded by the National Science and Technology Major Project (No.2023ZD0121103).
We would like to express gratitude to the anonymous reviewers for their kind comments.

%% file: 8_appendix.tex
\section{Optimization Details}\label{A}
In previous work~\citep{DBLP:conf/nips/MengBAB22, DBLP:conf/iclr/MengSABB23}, the optimal vector \( \mathbf{k}_* \) is derived using the following equation:
\begin{equation}
\mathbf{k}_* = \frac{1}{N} \sum_{j=1}^{N} \mathbf{k}(x_j + s),
\quad \text{where} \quad \mathbf{k}(x) = \sigma \left( W_{f_c}^{(l^*)} \gamma \left( a_{[x],i}^{(l^*)} + h_{[x],i}^{(l^*-1)} \right) \right).
\end{equation}
where \( \sigma \) represents the non-linearity applied to the hidden layer activations. Here, \( W_{f_c}^{(l^*)} \) represents the feed-forward weights of the layer being edited, and \( \gamma \) represents layer normalization applied to the activations from earlier layers.
For the value vector, we aim to optimize \( \mathbf{v}_* \) such that it minimizes the loss function \( \mathcal{L}(z) \), which is defined as:
\begin{equation}
\mathcal{L}(z) = \frac{1}{N} \sum_{j=1}^{N} \left[ - \log \mathbb{P}_{G(m_i^{l^*}:=z)} \left[ o^* \mid x_j + p \right] + D_{KL} \left( \mathbb{P}_{G(m_i^{l^*}:=z)} \left[ x \mid p' \right] \parallel \mathbb{P}_{G} \left[ x \mid p' \right] \right) \right],
\end{equation}
where the first term encourages the model to predict the target object \( o^* \) given the subject \( s \), and the second term controls the model’s behavior for unrelated prompts, maintaining the integrity of the rest of the knowledge base.

\section{Experimental Setup}\label{B}

\subsection{Baseline Editing Methods}
In our experiments, we selected four widely used knowledge editing methods as baselines, including:
\begin{itemize}
\item[$\bullet$] \textbf{ROME}~\citep{DBLP:conf/nips/MengBAB22}\footnote{https://github.com/kmeng01/rome}: ROME first identifies the factual knowledge at a specific layer within the transformer's MLP modules and then updates it by directly inserting new key-value pairs into the MLP module.
\item[$\bullet$]
\textbf{MEMIT}~\citep{DBLP:conf/iclr/MengSABB23}\footnote{https://github.com/kmeng01/memit}: Building upon ROME, MEMIT enables the editing of a larger collection of facts and performs updates across a sequence of MLP layers to modify the knowledge.
\item[$\bullet$] 
\textbf{RECT}~\citep{DBLP:conf/emnlp/GuXMLLCP24}\footnote{https://github.com/JasonForJoy/Model-Editing-Hurt}: RECT is designed to mitigate unintended negative impacts of knowledge editing on the broader capabilities of LLMs. Although editing can improve factual precision, it often compromises performance on tasks such as reasoning and question answering. RECT addresses this issue by regularizing weight modifications during editing, thereby restricting drastic changes that might lead to overfitting. Consequently, RECT achieves solid editing performance while maintaining the model’s overall general abilities.
\item[$\bullet$] 
\textbf{EMMET}~\citep{DBLP:conf/emnlp/GuptaBA24}\footnote{https://github.com/scalable-model-editing/unified-model-editing}: EMMET is a batched model editing method that unifies ROME and MEMIT under a single optimization framework. It enforces strict equality constraints to precisely update model weights while preserving existing knowledge, enabling large-scale edits. By stabilizing matrix computations and decoupling edit distribution from optimization, EMMET achieves performance matching MEMIT while revealing the fundamental equivalence between these approaches. The method maintains model quality across key metrics, offering an efficient solution for mass-editing transformers.
\item[$\bullet$] 
\textbf{PRUNE}~\citep{DBLP:conf/iclr/MaWXLG25}\footnote{https://github.com/mjy1111/PRUNE}: PRUNE is a model editing framework that maintains the overall capabilities of large language models even as multiple edits are applied. To prevent performance degradation from accumulating edits, PRUNE imposes a constraint on the condition number of the editing matrix. This constraint helps minimize interference with the model’s existing knowledge. By regulating the model’s numerical stability, PRUNE ensures that its general performance remains robust throughout the editing process.
\item[$\bullet$] 
\textbf{Alpha-Edit}~\citep{DBLP:conf/iclr/FangJWMSW0C25}\footnote{https://github.com/ jianghoucheng/AlphaEdit}: Alpha-Edit combats catastrophic forgetting in knowledge editing by safeguarding previously acquired knowledge and extends this capability into a lifelong editing scenario through a null-space projection technique. This approach ensures that new updates minimally interfere with prior knowledge by projecting edits into directions that have little impact on existing representations, thus preserving model stability even across numerous edits.
\end{itemize}
The effectiveness of these methods was evaluated using EasyEdit~\citep{DBLP:journals/corr/abs-2308-07269}, a user-friendly knowledge editing framework that consolidates publicly available code and hyperparameters from prior approaches.

\subsubsection{Editing Datasets}
\begin{figure*}[t]
  \centering
  \includegraphics[width=0.7\textwidth]{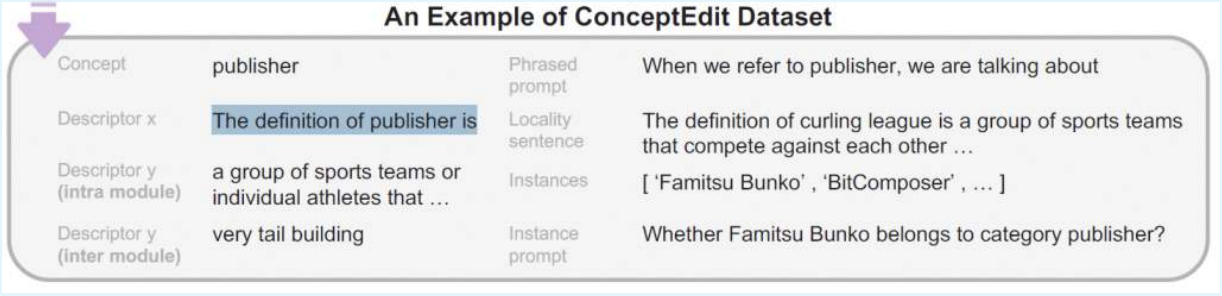}
  \vspace{-4mm}
  \caption{An example of ConceptEdit dataset.}
  \vspace{-3mm}
  \label{conceptedit}
\end{figure*}

In our experiments, we utilized two widely used factual knowledge editing datasets, \textsc{ZsRE}~\citep{DBLP:conf/conll/LevySCZ17} and \textsc{CounterFact}~\citep{DBLP:conf/nips/MengBAB22}. Examples from these two datasets are shown in Table~\ref{zsre-counterfact}. Additionally, we included a conceptual knowledge editing dataset named \textsc{ConceptEdit}~\citep{DBLP:conf/emnlp/WangMDYSLGC024}, with examples illustrated in Figure~\ref{conceptedit}.

\textbf{\textsc{ZsRE}} is a question-answering dataset that employs back-translation to generate paraphrased questions as equivalent variations. Each sample consists of a question related to an entity, and alternative plausible edit labels are drawn from the top-ranked outputs of a BART-base model trained specifically on \textsc{ZsRE}.

\textbf{\textsc{CounterFact}} focuses on counterfactual knowledge that initially receives lower confidence scores compared to established facts. It generates out-of-distribution examples by replacing the subject entity with a closely related entity sharing the same predicate. This modification helps distinguish between superficial textual changes and more substantial edits that reflect a meaningful and genuine shift in the underlying factual knowledge.

\begin{table*}[h]
\centering
\caption{The editing datasets of both \textsc{ZsRE} and \textsc{CounterFact}.}
\begin{tabular}{ll}
\toprule
\textbf{Datasets} & \textbf{Editing prompt} \\
\midrule
\textsc{ZsRE} & Which was the record label for New Faces, New Sounds? \\
\textsc{CounterFact} & In America, the official language is \\
\bottomrule
\end{tabular}
\label{zsre-counterfact}
\end{table*}

\textbf{\textsc{ConceptEdit}} targets the modification of conceptual knowledge instead of isolated factual statements. It introduces a benchmark based on the DBpedia Ontology, where both concept definitions and their linked instances are altered to examine how changes spread within the model. This framework allows for distinguishing between minor surface-level edits and significant revisions that indicate a real shift in the model’s comprehension of abstract concepts.

\subsection{Evaluation Metrics}
\paragraph{Metrics for Evaluating Editing Performance}
We assess editing performance using three key metrics: \textbf{\textit{reliability}}, \textbf{\textit{generalization}}, and \textbf{\textit{locality}}~\citep{DBLP:conf/emnlp/YaoWT0LDC023}.
\textbf{\textit{Reliability}} refers to how accurately the edited model produces the intended predictions when queried with the modified knowledge. It is quantified as the mean accuracy over the edited examples:
\begin{equation}
    \mathrm{E}_{(x'_{ei}, y'_{ei}) \sim \{(x_{ei}, y_{ei})\}} \mathbf{1} \left\{ \arg\max_y f_{\theta_{i}} \left( y \mid x'_{ei} \right) = y'_{ei} \right\}.
\end{equation}
\textbf{\textit{Generalization}} indicates the model’s capacity to recall and apply the newly updated knowledge when presented with inputs related to the editing context. It is evaluated by calculating the average accuracy on examples drawn uniformly from the equivalence neighborhood:
\begin{equation}
    \mathrm{E}_{(x'_{ei}, y'_{ei}) \sim N(x_{ei}, y_{ei})} \mathbf{1} \left\{ \arg\max_y f_{\theta_{i}} \left( y \mid x'_{ei} \right) = y'_{ei} \right\}.
\end{equation}
\textbf{\textit{Locality}} measures the extent to which the model’s behavior remains stable on inputs unrelated to the edited knowledge. Locality is calculated as the proportion of outputs that remain consistent between the edited model and its pre-edit version for out-of-scope prompts:
\begin{equation}
    \mathrm{E}_{(x'_{ei}, y'_{ei}) \sim O(x_{ei}, y_{ei})} \mathbf{1} \left\{ f_{\theta_{i}} \left( y \mid x'_{ei} \right) = f_{\theta_{i-1}} \left( y \mid x'_{ei} \right) \right\}.
\end{equation}

\paragraph{Metrics for Evaluating General Abilities}
Four downstream tasks were chosen to evaluate the general abilities of models before and after editing:
\textbf{\textit{Natural language inference (NLI)}} was evaluated on the RTE~\citep{DBLP:conf/mlcw/DaganGM05} via binary classification accuracy.
\textbf{\textit{Open-domain QA}} was evaluated on the Natural Questions~\citep{DBLP:journals/tacl/KwiatkowskiPRCP19}, where results were calculated using exact match (EM) scores against reference answers, following minor normalization as in \citep{DBLP:conf/acl/ChenFWB17} and \citep{DBLP:conf/acl/LeeCT19}.
\textbf{\textit{Summarization}} utilized the SAMSum dataset~\citep{gliwa-etal-2019-samsum}, with evaluation based on the average of ROUGE-1, ROUGE-2, and ROUGE-L scores, consistent with~\citep{lin-2004-rouge}.
\textbf{\textit{Sentiment analysis}} was performed on SST2~\citep{DBLP:conf/emnlp/SocherPWCMNP13}, measured by binary classification accuracy.

\subsubsection{Downstream Task Prompts}
The prompts for each task were illustrated in Table~\ref{tab-prompt}.

\begin{table}[h]
% \small
\centering
\caption{The prompts to LLMs for evaluating their zero-shot performance on these general tasks.}
\begin{tabular}{p{0.95\linewidth}}
\toprule
NLI:\\
\{\texttt{SENTENCE1}\} entails the \{\texttt{SENTENCE2}\}. True or False? answer:\\
\midrule
Open-domain QA:\\
Refer to the passage below and answer the following question. Passage: \{\texttt{DOCUMENT}\} Question: \{\texttt{QUESTION}\}\\
\midrule
Summarization:\\
\{\texttt{DIALOGUE}\} TL;DR:\\
\midrule
Sentiment analysis:\\
For each snippet of text, label the sentiment of the text as positive or negative. The answer should be exact 'positive' or 'negative'. text: \{\texttt{TEXT}\} answer:\\
\bottomrule
\end{tabular}
\label{tab-prompt}
\end{table}

\section{Experimental Results}\label{C}

\subsection{Results in Single-Sequential Editing Scenarios}
\paragraph{Editing Performance}
Table~\ref{single-sequential-t1} and Table~\ref{single-sequential-t-other} illustrate the editing performance of models based on LLaMA3-8B and Qwen2.5-7B on four different editing datasets.
Table~\ref{single-sequential-t-llama2} further shows the editing performance of models based on LLaMA2-13B on four different editing datasets. Overall, the results from these experiments are consistent with our earlier observations and provide strong evidence for the effectiveness of \METHODNAME{} in maintaining editing performance.

\begin{table*}
    \centering
    \caption{In the single-sequential editing scenario, the editing performance of different methods on CounterFact and ConceptEdit-Inter. We performed 4000 sequential edits.}
    \label{single-sequential-t1}
    \small
    \renewcommand{\arraystretch}{0.9}
    \setlength{\tabcolsep}{4pt}
    \resizebox{\linewidth}{!}{\begin{tabular}{lccccccc}
        \toprule
        \multirow{2}{*}{Method} & \multirow{2}{*}{Model} & \multicolumn{3}{c}{\textbf{CounterFact}} & \multicolumn{3}{c}{\textbf{ConceptEdit-Inter}}\\
        \cmidrule(lr){3-5} \cmidrule(lr){6-8}
        & & Reliability & Generalization & Locality & Reliability & Generalization & Locality \\
        \midrule
        ROME & \multirow{6}{*}{LLaMA3-8B} & 0.0000 & 0.0000 & 0.0000 & 0.0000 & 0.0000 & 0.0000 \\
        MEMIT & & 0.0000 & 0.0000 & 0.0000 & 0.0000 & 0.0000 & 0.0000 \\
        RECT & & 0.5688 & 0.3288 & 0.2517 & 0.3531 & 0.2084 & 0.1502 \\
        EMMET & & 0.6398 & 0.4937 & 0.3253 & 0.3877 & 0.2258 & 0.1724 \\
        PRUNE & & 0.7886 & 0.7140 & 0.5763 & 0.6295 & 0.4414 & 0.3193 \\
        AlphaEdit & & 0.9018 & 0.8260 & 0.7831 & 0.7012 & 0.6007 & 0.5212 \\
        \midrule
        \METHODNAME{} & & \textbf{0.9887} & \textbf{0.9863} & \textbf{0.8972} & \textbf{0.7859} & \textbf{0.7275} & \textbf{0.6856} \\
        \midrule
        \midrule
        ROME & \multirow{6}{*}{Qwen2.5-7B} & 0.0000 & 0.0000 & 0.0000 & 0.0000 & 0.0000 & 0.0000 \\
        MEMIT & & 0.0000 & 0.0000 & 0.0000 & 0.0000 & 0.0000 & 0.0000 \\
        RECT & & 0.6203 & 0.4745 & 0.3582 & 0.3737 & 0.2306 & 0.1738 \\
        EMMET & & 0.6702 & 0.5589 & 0.4771 & 0.4593 & 0.2641 & 0.1903 \\
        PRUNE & & 0.8115 & 0.7860 & 0.6823 & 0.6708 & 0.5009 & 0.4120 \\
        AlphaEdit & & 0.9519 & 0.9241 & 0.8418 & 0.7346 & 0.6453 & 0.6116 \\
        \midrule
        \METHODNAME{} & & \textbf{0.9981} & \textbf{0.9902} & \textbf{0.9098} & \textbf{0.8012} & \textbf{0.7547} & \textbf{0.7069} \\
        \midrule
        \bottomrule
    \end{tabular}
    }
\end{table*}

\begin{table*}
    \centering
    \caption{In the single-sequential editing scenario, the editing performance of different methods on ZsRE and ConceptEdit-Intra. We performed 4000 sequential edits.}
    \small
    \renewcommand{\arraystretch}{0.9}
    \setlength{\tabcolsep}{4pt}
    \label{single-sequential-t-other}
    \resizebox{\linewidth}{!}{
    \begin{tabular}{lccccccc}
        \toprule
        \multirow{2}{*}{Method} & \multirow{2}{*}{Model} & \multicolumn{3}{c}{\textbf{ZsRE}} & \multicolumn{3}{c}{\textbf{ConceptEdit-Intra}}\\
        \cmidrule(lr){3-5} \cmidrule(lr){6-8}
        & & Reliability & Generalization & Locality & Reliability & Generalization & Locality \\
        \midrule
        ROME & \multirow{6}{*}{LLaMA3-8B} & 0.0000 & 0.0000 & 0.0000 & 0.0000 & 0.0000 & 0.0000 \\
        MEMIT & & 0.0000 & 0.0000 & 0.0000 & 0.0000 & 0.0000 & 0.0000 \\
        RECT & & 0.6029 & 0.3894 & 0.2922 & 0.4012 & 0.2756 & 0.2177 \\
        EMMET & & 0.6902 & 0.5311 & 0.3544 & 0.4226 & 0.2446 & 0.1820 \\
        PRUNE & & 0.8506 & 0.7727 & 0.6255 & 0.6771 & 0.4764 & 0.3496 \\
        AlphaEdit & & 0.9675 & 0.8932 & 0.8113 & 0.8145 & 0.7192 & 0.6335 \\
        \midrule
        \METHODNAME{} & & \textbf{0.9962} & \textbf{0.9971} & \textbf{0.8978} & \textbf{0.9698} & \textbf{0.8375} & \textbf{0.7102} \\
        \midrule
        \midrule
        ROME & \multirow{6}{*}{Qwen2.5-7B} & 0.0000 & 0.0000 & 0.0000 & 0.0000 & 0.0000 & 0.0000 \\
        MEMIT & & 0.0000 & 0.0000 & 0.0000 & 0.0000 & 0.0000 & 0.0000 \\
        RECT & & 0.6327 & 0.4108 & 0.3286 & 0.4283 & 0.2981 & 0.2692 \\
        EMMET & & 0.6880 & 0.5417 & 0.3566 & 0.4143 & 0.2414 & 0.1814 \\
        PRUNE & & 0.8518 & 0.7685 & 0.6153 & 0.6783 & 0.4658 & 0.3402 \\
        AlphaEdit & & 0.9644 & 0.9153 & 0.8369 & 0.8476 & 0.7658 & 0.6477 \\
        \midrule
        \METHODNAME{} & & \textbf{0.9987} & \textbf{0.9985} & \textbf{0.9237} & \textbf{0.9752} & \textbf{0.8530} & \textbf{0.7474} \\
        \midrule
        \bottomrule
    \end{tabular}
    }
\end{table*}

\begin{table*}
    \centering
    \caption{In the single-sequential editing scenario, the editing performance of different methods on CounterFact, ZsRE, ConceptEdit-intra and ConceptEdit-inter. We performed 4000 sequential edits.}
    \small
    \renewcommand{\arraystretch}{0.9}
    \setlength{\tabcolsep}{4pt}
    \label{single-sequential-t-llama2}
    \resizebox{\linewidth}{!}{
    \begin{tabular}{lccccccc}
        \toprule
        \multirow{2}{*}{Method} & \multirow{2}{*}{Model} & \multicolumn{3}{c}{\textbf{CounterFact}} & \multicolumn{3}{c}{\textbf{ConceptEdit-Inter}}\\
        \cmidrule(lr){3-5} \cmidrule(lr){6-8}
        & & Reliability & Generalization & Locality & Reliability & Generalization & Locality \\
        \midrule
        ROME & \multirow{6}{*}{LLaMA2-13B} & 0.0000 & 0.0000 & 0.0000 & 0.0000 & 0.0000 & 0.0000 \\
        MEMIT & & 0.0000 & 0.0000 & 0.0000 & 0.0000 & 0.0000 & 0.0000 \\
        RECT & & 0.5294 & 0.3093 & 0.2378 & 0.3353 & 0.1971 & 0.1423 \\
        EMMET &  & 0.6732 & 0.5147 & 0.3345 & 0.4211 & 0.2343 & 0.1695 \\
        PRUNE &  & 0.8415 & 0.7502 & 0.5627 & 0.6154 & 0.4396 & 0.3339 \\
        AlphaEdit & & 0.8445 & 0.7714 & 0.7402 & 0.6613 & 0.5690 & 0.4902 \\
        \midrule
        \METHODNAME{} & & \textbf{0.9250} & \textbf{0.9293} & \textbf{0.8500} & \textbf{0.7441} & \textbf{0.6781} & \textbf{0.6497} \\
        \midrule
        \midrule
        \multirow{2}{*}{Method} & \multirow{2}{*}{Model} & \multicolumn{3}{c}{\textbf{ZsRE}} & \multicolumn{3}{c}{\textbf{ConceptEdit-Intra}}\\
        \cmidrule(lr){3-5} \cmidrule(lr){6-8}
        & & Reliability & Generalization & Locality & Reliability & Generalization & Locality \\
        \midrule
        ROME & \multirow{6}{*}{LLaMA2-13B} & 0.0000 & 0.0000 & 0.0000 & 0.0000 & 0.0000 & 0.0000 \\
        MEMIT & & 0.0000 & 0.0000 & 0.0000 & 0.0000 & 0.0000 & 0.0000 \\
        RECT & & 0.5886 & 0.3864 & 0.3098 & 0.4019 & 0.2793 & 0.2545 \\
        EMMET & & 0.6903 & 0.5416 & 0.3541 & 0.4244 & 0.2431 & 0.1853 \\
        PRUNE & & 0.8489 & 0.7710 & 0.6223 & 0.6789 & 0.4762 & 0.3432 \\
        AlphaEdit & & 0.9157 & 0.8602 & 0.7857 & 0.8021 & 0.7145 & 0.6024 \\
        \midrule
        \METHODNAME{} & & \textbf{0.9471} & \textbf{0.9289} & \textbf{0.8635} & \textbf{0.9157} & \textbf{0.8094} & \textbf{0.6992} \\
        \midrule
        \bottomrule
    \end{tabular}
    }
\end{table*}

\begin{figure*}[t]
    \centering
    \includegraphics[width=0.9\linewidth]{figures/legend_new.pdf}
    \includegraphics[width=1\linewidth]{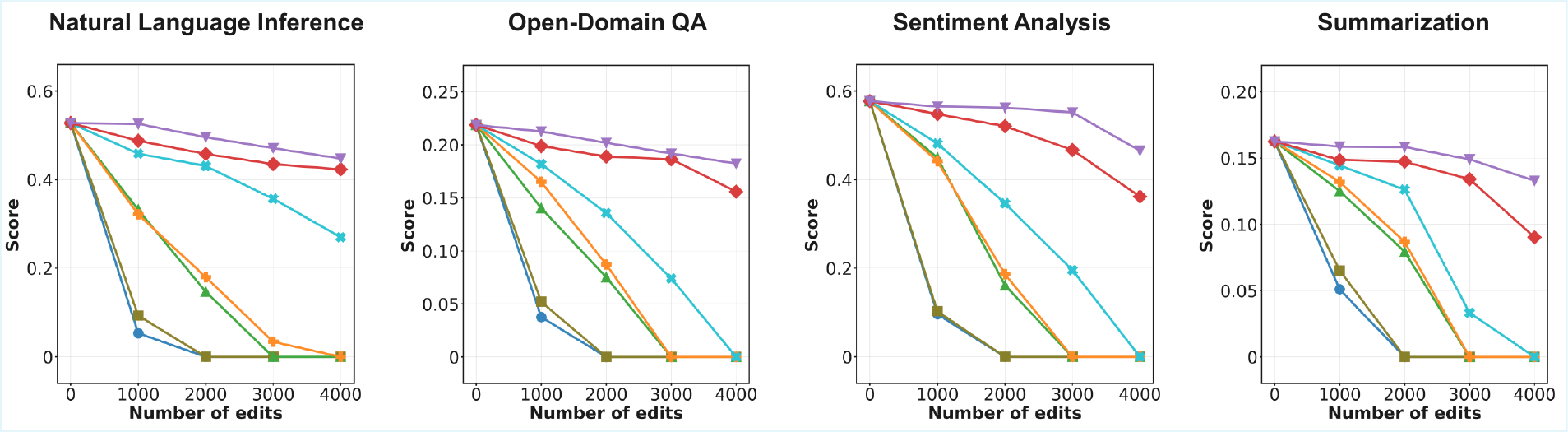}
    \captionof{figure}{Edited on the CounterFact dataset, the general task performance of varying methods with LLaMA2-13B as the number of edits increases, under the single-sequential scenario.}
    \label{single-sequential-f-ct-llama2}
\end{figure*}

\begin{figure*}[t]
    \centering
    \includegraphics[width=0.9\linewidth]{figures/legend_new.pdf}
    \includegraphics[width=1\linewidth]{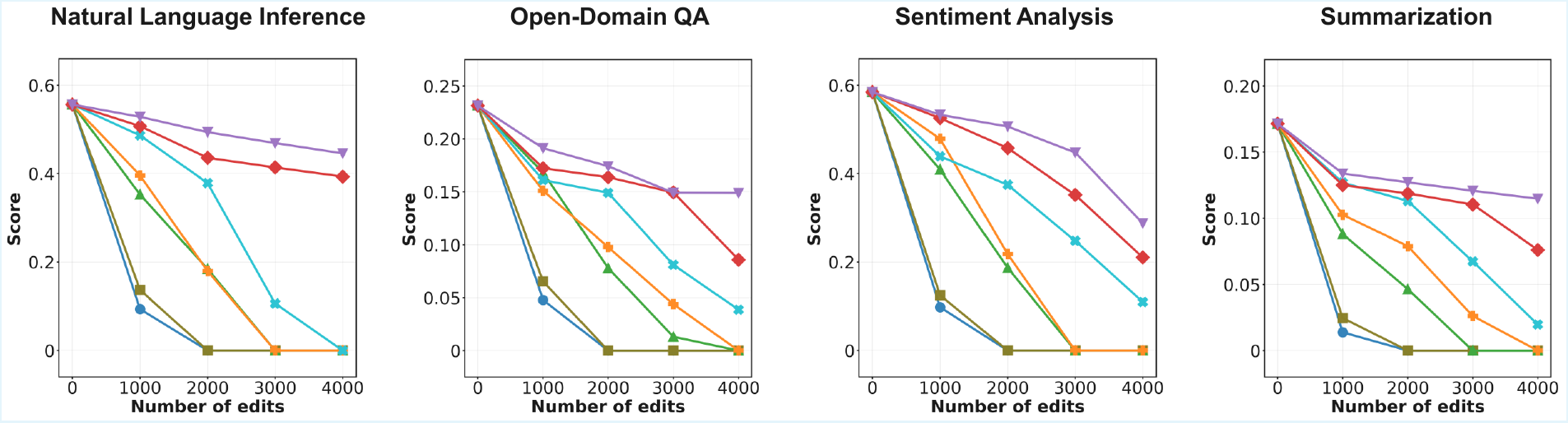}
    \caption{Edited on the CounterFact dataset, the general task performance of varying methods with Qwen2.5-7B as the number of edits increases, under the single-sequential scenario.}
    \label{single-sequential-f-ct-qwen}
\end{figure*}

\begin{figure*}[t]
    \centering
    \includegraphics[width=0.9\linewidth]{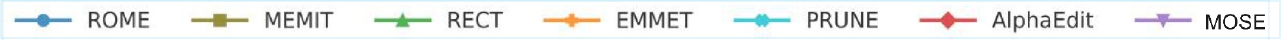}
    \includegraphics[width=1\linewidth]{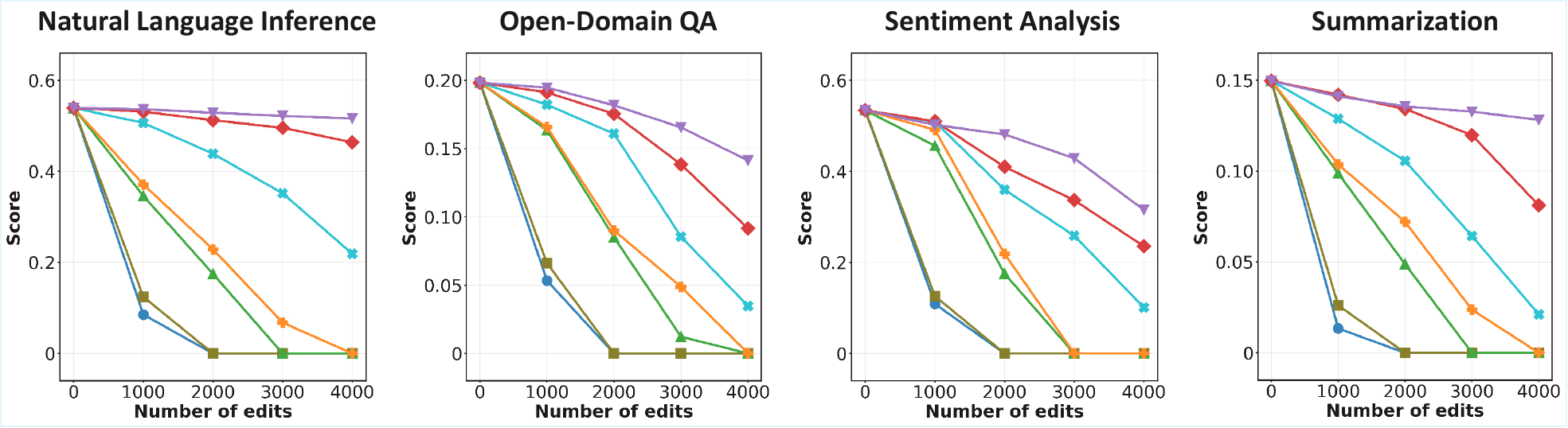}
    \captionof{figure}{Edited on the CounterFact dataset, the general task performance of varying methods with LLaMA3-8B as the number of edits increases, under the single-sequential scenario.}
    \label{single-sequential-f1}
\end{figure*}

\begin{figure*}[t]
    \centering
    \includegraphics[width=0.9\linewidth]{figures/legend_new.pdf}
    \includegraphics[width=0.95\linewidth]{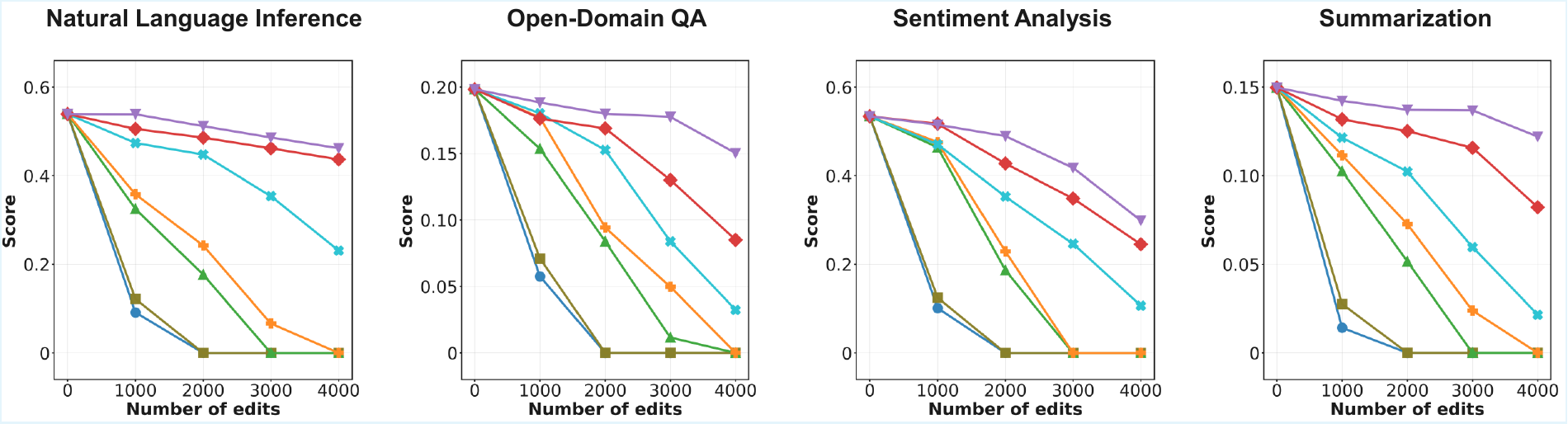}
    \caption{Edited on the ZsRE dataset, the general task performance of varying methods with LLaMA3-8B as the number of edits increases, under the single-sequential scenario.}
    \label{single-sequential-f-zsre-llama3}
\end{figure*}

\begin{figure*}[t]
    \centering
    \includegraphics[width=0.9\linewidth]{figures/legend_new.pdf}
    \includegraphics[width=0.95\linewidth]{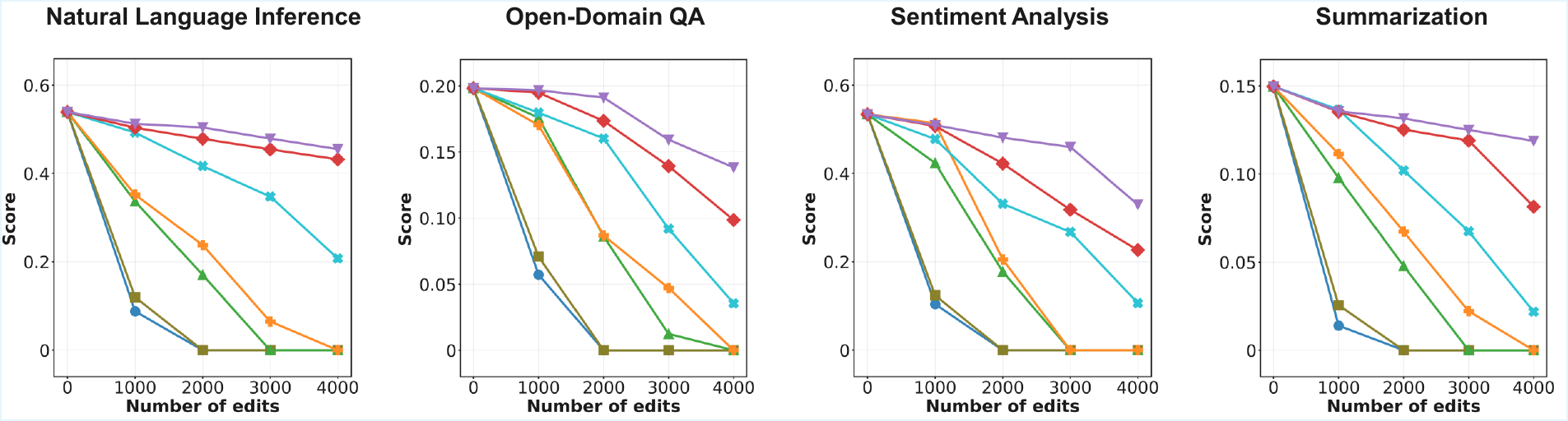}
    \caption{Edited on the ConceptEdit-Intra dataset, the general task performance of varying methods with LLaMA3-8B as the number of edits increases, under the single-sequential scenario.}
    \label{single-sequential-f-intra-llama3}
\end{figure*}

\begin{figure*}[t]
    \centering
    \includegraphics[width=0.9\linewidth]{figures/legend_new.pdf}
    \includegraphics[width=0.95\linewidth]{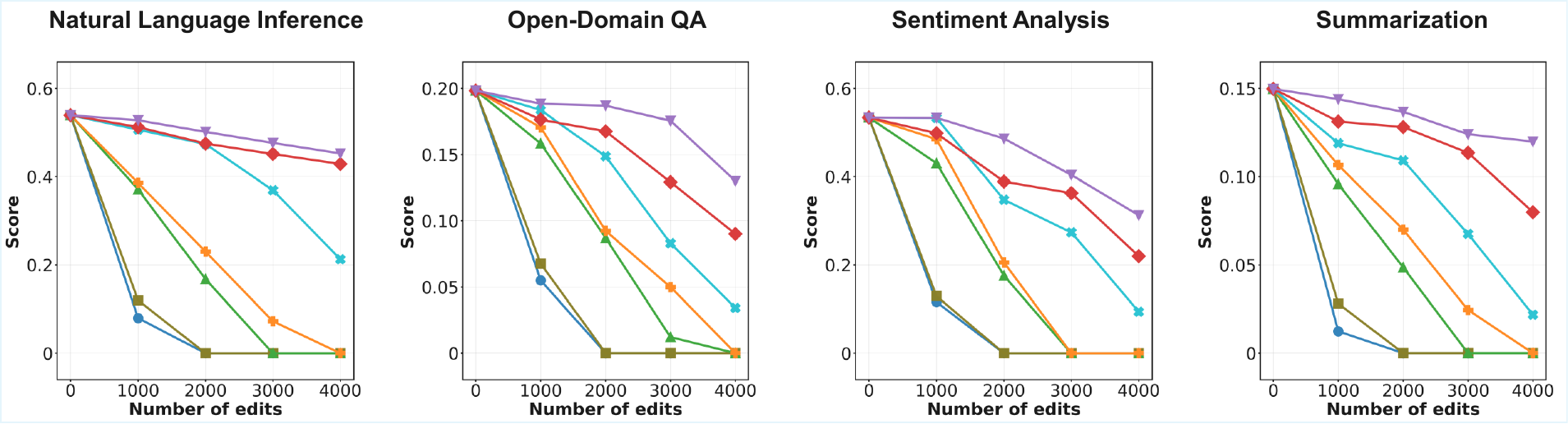}
    \caption{Edited on the ConceptEdit-Inter dataset, the general task performance of varying methods with LLaMA3-8B as the number of edits increases, under the single-sequential scenario.}
    \label{single-sequential-f-inter-llama3}
\end{figure*}

\begin{figure*}[t]
    \centering
    \includegraphics[width=0.9\linewidth]{figures/legend_new.pdf}
    \includegraphics[width=0.95\linewidth]{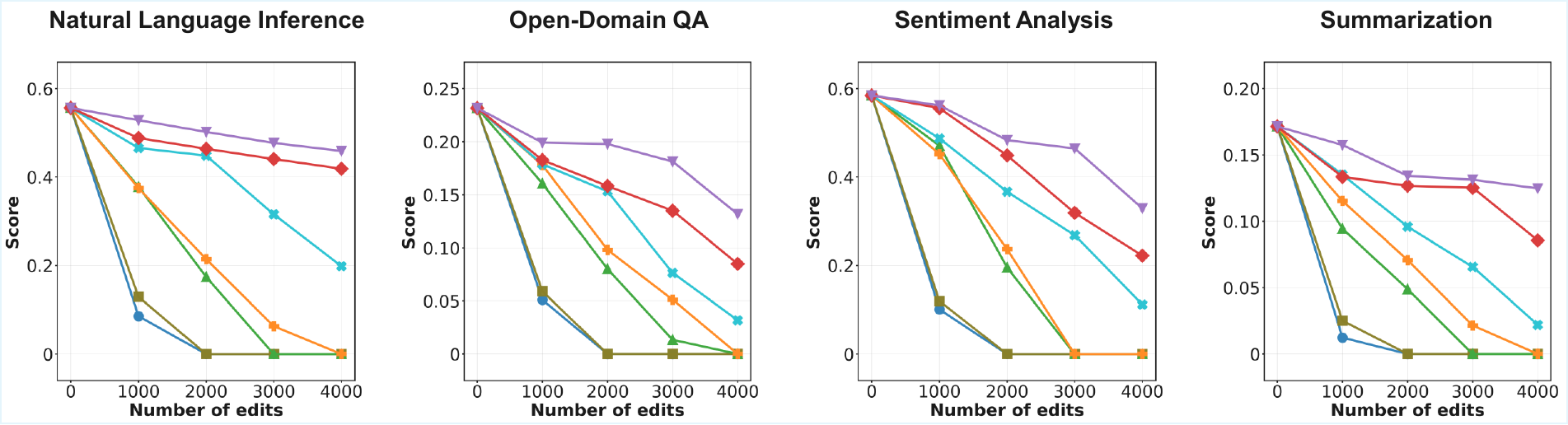}
    \caption{Edited on the ConceptEdit-Inter dataset, the general task performance of varying methods with Qwen2.5-7B as the number of edits increases, under the single-sequential scenario.}
    \label{single-sequential-f-inter-qwen}
\end{figure*}

\begin{figure*}[t]
    \centering
    \includegraphics[width=0.9\linewidth]{figures/legend_new.pdf}
    \includegraphics[width=0.95\linewidth]{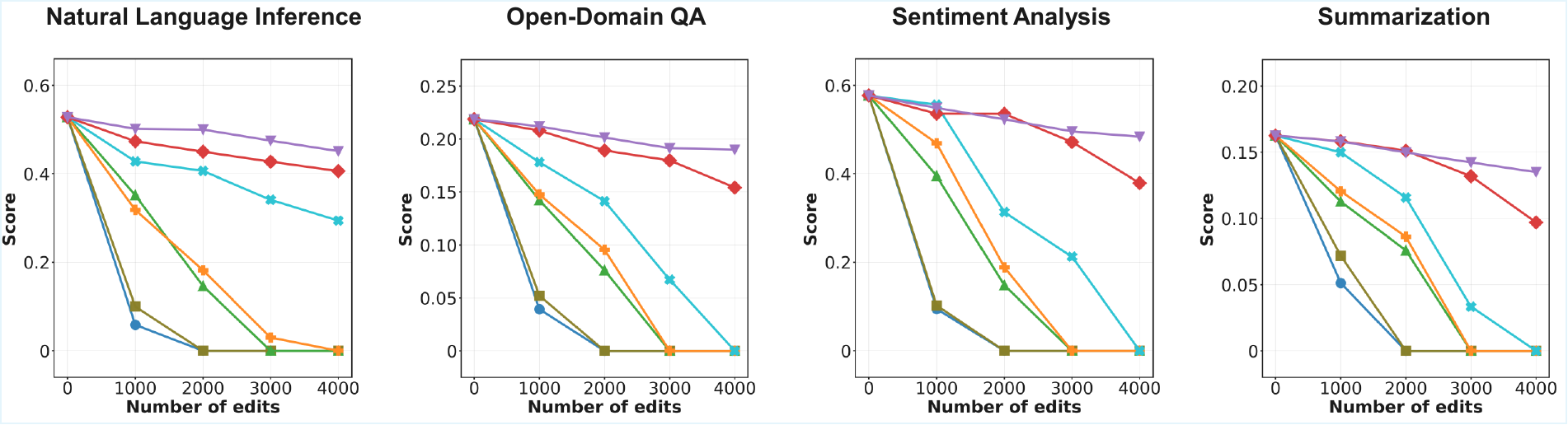}
    \caption{Edited on the ConceptEdit-Inter dataset, the general task performance of varying methods with LLaMA2-13B as the number of edits increases, under the single-sequential scenario.}
    \label{single-sequential-f-inter-llama2}
\end{figure*}

\paragraph{General Abilities}
Figure~\ref{single-sequential-f-ct-llama2} and Figure~\ref{single-sequential-f-ct-qwen} illustrate the general abilities of models based on LLaMA2-13B and Qwen2.5-7B on the CounterFact dataset. Figure~\ref{single-sequential-f1}, Figure~\ref{single-sequential-f-zsre-llama3}, Figure~\ref{single-sequential-f-intra-llama3} and Figure~\ref{single-sequential-f-inter-llama3} further show the general abilities of the model using LLaMA3-8B on the CounterFact, ZsRE, ConceptEdit-Intra and ConceptEdit-Inter. 
Similarly, Figure~\ref{single-sequential-f-inter-qwen} illustrates the general abilities of the Qwen2.5-7B ConceptEdit-Inter datasets, while Figure~\ref{single-sequential-f-inter-llama2} illustrates the general abilities of the LLaMA2-13B on the same dataset. 
Overall, the results from these experiments are consistent with our earlier observations and provide strong evidence for the effectiveness of \METHODNAME{} in maintaining general abilities.

\subsection{Results in Batch-Sequential Editing Scenarios}
\paragraph{Editing Performance}
Table~\ref{batch-sequential-t-other} illustrates the editing performance of models based on LLaMA3-8B and Qwen2.5-7B on the ZsRE and ConceptEdit-Intra. Table~\ref{batch-sequential-t-llama2} further shows the editing performance of models based on LLaMA2-13B on four different editing datasets. Overall, the results from these experiments are consistent with our earlier observations and provide strong evidence for the effectiveness of \METHODNAME{} in maintaining editing performance. In our experiments, we set the batch size to 100 and performed 100 sequential edits.
Overall, the results are consistent with our earlier observations and provide strong evidence for the effectiveness of \METHODNAME{} in maintaining editing performance, even in more challenging scenarios.

\begin{table*}
    \centering
    \caption{In the batch-sequential editing scenario, the editing performance of different methods on ZsRE and ConceptEdit-Intra. We set the batch size to 100 and performed 100 sequential edits.}
    \small
    \renewcommand{\arraystretch}{0.9}
    \setlength{\tabcolsep}{4pt}
    \label{batch-sequential-t-other}
    \resizebox{\linewidth}{!}{
    \begin{tabular}{lccccccc}
        \toprule
        \multirow{2}{*}{Method} & \multirow{2}{*}{Model} & \multicolumn{3}{c}{\textbf{ZsRE}} & \multicolumn{3}{c}{\textbf{ConceptEdit-Intra}}\\
        \cmidrule(lr){3-5} \cmidrule(lr){6-8}
        & & Reliability & Generalization & Locality & Reliability & Generalization & Locality \\
        \midrule
        ROME & \multirow{6}{*}{LLaMA3-8B} & 0.0000 & 0.0000 & 0.0000 & 0.0000 & 0.0000 & 0.0000 \\
        MEMIT & & 0.0000 & 0.0000 & 0.0000 & 0.0000 & 0.0000 & 0.0000 \\
        RECT &  & 0.5419 & 0.3182 & 0.2423 & 0.3218 & 0.1991 & 0.1430 \\
        EMMET &  & 0.6394 & 0.4916 & 0.3181 & 0.4010 & 0.2224 & 0.1597 \\
        PRUNE &  & 0.7954 & 0.7072 & 0.5336 & 0.5825 & 0.4204 & 0.3150 \\
        AlphaEdit &  & 0.8453 & 0.7841 & 0.7036 & 0.7197 & 0.6003 & 0.5022 \\
        \midrule
        \METHODNAME{} & & \textbf{0.9420} & \textbf{0.9510} & \textbf{0.9137} & \textbf{0.7592} & \textbf{0.7222} & \textbf{0.6640} \\
        \midrule
        \midrule
        ROME & \multirow{6}{*}{Qwen2.5-7B} & 0.0000 & 0.0000 & 0.0000 & 0.0000 & 0.0000 & 0.0000 \\
        MEMIT & & 0.0000 & 0.0000 & 0.0000 & 0.0000 & 0.0000 & 0.0000 \\
        RECT &  & 0.6166 & 0.4741 & 0.3276 & 0.3668 & 0.2260 & 0.1581 \\
        EMMET &  & 0.6398 & 0.4849 & 0.3182 & 0.3973 & 0.2241 & 0.1591 \\
        PRUNE &  & 0.7923 & 0.7088 & 0.5300 & 0.5783 & 0.4174 & 0.3127 \\
        AlphaEdit &  & 0.9223 & 0.8995 & 0.8268 & 0.6861 & 0.6253 & 0.5712 \\
        \midrule
        \METHODNAME{} & & \textbf{0.9421} & \textbf{0.9443} & \textbf{0.8941} & \textbf{0.7728} & \textbf{0.7170} & \textbf{0.6841} \\
        \midrule
        \bottomrule
    \end{tabular}
    }
\end{table*}

\begin{table*}
    \centering
    \caption{In the batch-sequential editing scenario, the editing performance of different methods on CounterFact, ZsRE, ConceptEdit-Intra and ConceptEdit-Inter. We set the batch size to 100 and performed 100 sequential edits.}
    \small
    \renewcommand{\arraystretch}{0.9}
    \setlength{\tabcolsep}{4pt}
    \label{batch-sequential-t-llama2}
    \resizebox{\linewidth}{!}{
    \begin{tabular}{lccccccc}
        \toprule
        \multirow{2}{*}{Method} & \multirow{2}{*}{Model} & \multicolumn{3}{c}{\textbf{CounterFact}} & \multicolumn{3}{c}{\textbf{ConceptEdit-Inter}}\\
        \cmidrule(lr){3-5} \cmidrule(lr){6-8}
        & & Reliability & Generalization & Locality & Reliability & Generalization & Locality \\
        \midrule
        ROME & \multirow{6}{*}{LLaMA2-13B} & 0.0000 & 0.0000 & 0.0000 & 0.0000 & 0.0000 & 0.0000 \\
        MEMIT & & 0.0000 & 0.0000 & 0.0000 & 0.0000 & 0.0000 & 0.0000 \\
        RECT &  & 0.4661 & 0.2757 & 0.2115 & 0.2897 & 0.1763 & 0.1241 \\
        EMMET &  & 0.5388 & 0.4987 & 0.4204 & 0.4285 & 0.2380 & 0.1712 \\
        PRUNE &  & 0.6557 & 0.6268 & 0.6082 & 0.5265 & 0.4617 & 0.3736 \\
        AlphaEdit &  & 0.7283 & 0.6975 & 0.6315 & 0.6202 & 0.5215 & 0.4378 \\
        \midrule
        \METHODNAME{} & & \textbf{0.8432} & \textbf{0.8477} & \textbf{0.7734} & \textbf{0.6765} & \textbf{0.6263} & \textbf{0.5731} \\
        \midrule
        \midrule
        \multirow{2}{*}{Method} & \multirow{2}{*}{Model} & \multicolumn{3}{c}{\textbf{ZsRE}} & \multicolumn{3}{c}{\textbf{ConceptEdit-Intra}}\\
        \cmidrule(lr){3-5} \cmidrule(lr){6-8}
        & & Reliability & Generalization & Locality & Reliability & Generalization & Locality \\
        \midrule
        ROME & \multirow{6}{*}{LLaMA2-13B} & 0.0000 & 0.0000 & 0.0000 & 0.0000 & 0.0000 & 0.0000 \\
        MEMIT & & 0.0000 & 0.0000 & 0.0000 & 0.0000 & 0.0000 & 0.0000 \\
        RECT &  & 0.4902 & 0.2943 & 0.2200 & 0.3021 & 0.1823 & 0.1382 \\
        EMMET &  & 0.6370 & 0.5022 & 0.4235 & 0.4298 & 0.2415 & 0.1718 \\
        PRUNE &  & 0.7484 & 0.7310 & 0.6094 & 0.6186 & 0.4662 & 0.3724 \\
        AlphaEdit &  & 0.8741 & 0.8398 & 0.7471 & 0.6348 & 0.5885 & 0.5361 \\
        \midrule
        \METHODNAME{} & & \textbf{0.8748} & \textbf{0.8709} & \textbf{0.8302} & \textbf{0.7010} & \textbf{0.6963} & \textbf{0.6079} \\
        \midrule
        \bottomrule
    \end{tabular}
    }
\end{table*}

\begin{figure*}[t]
    \centering
    \includegraphics[width=0.9\linewidth]{figures/legend_new.pdf}
    \includegraphics[width=0.95\linewidth]{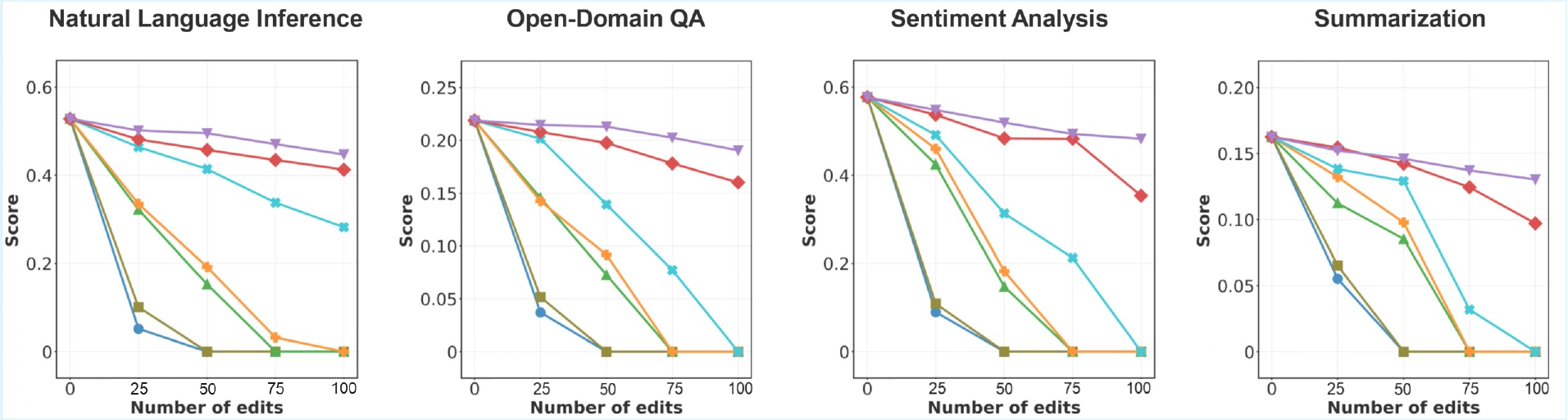}
    \caption{Edited on the CounterFact dataset, the general task performance of varying methods with LLaMA2-13B as the number of edits increases, under the batch-sequential scenario. We set the batch size to 100.}
    \label{batch-sequential-f-ct-llama2}
\end{figure*}

\begin{figure*}[t]
    \centering
    \includegraphics[width=0.9\linewidth]{figures/legend_new.pdf}
    \includegraphics[width=0.95\linewidth]{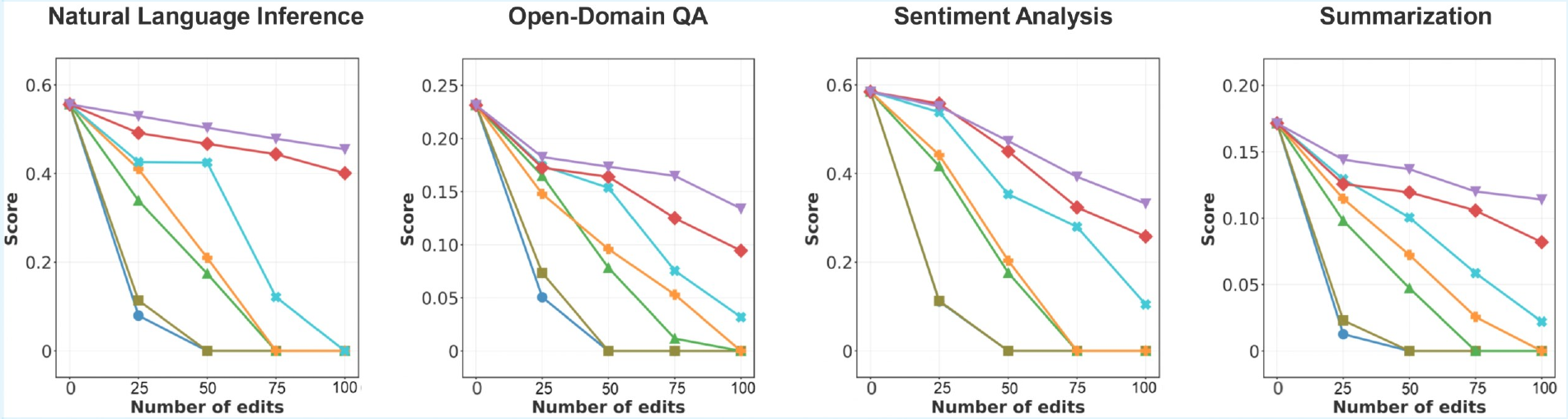}
    \caption{Edited on the CounterFact dataset, the general task performance of varying methods with Qwen2.5-7B as the number of edits increases, under the batch-sequential scenario. We set the batch size to 100.}
    \label{batch-sequential-f-ct-qwen}
\end{figure*}

\begin{figure*}[t]
    \centering
    \includegraphics[width=0.9\linewidth]{figures/legend_new.pdf}
    \includegraphics[width=0.95\linewidth]{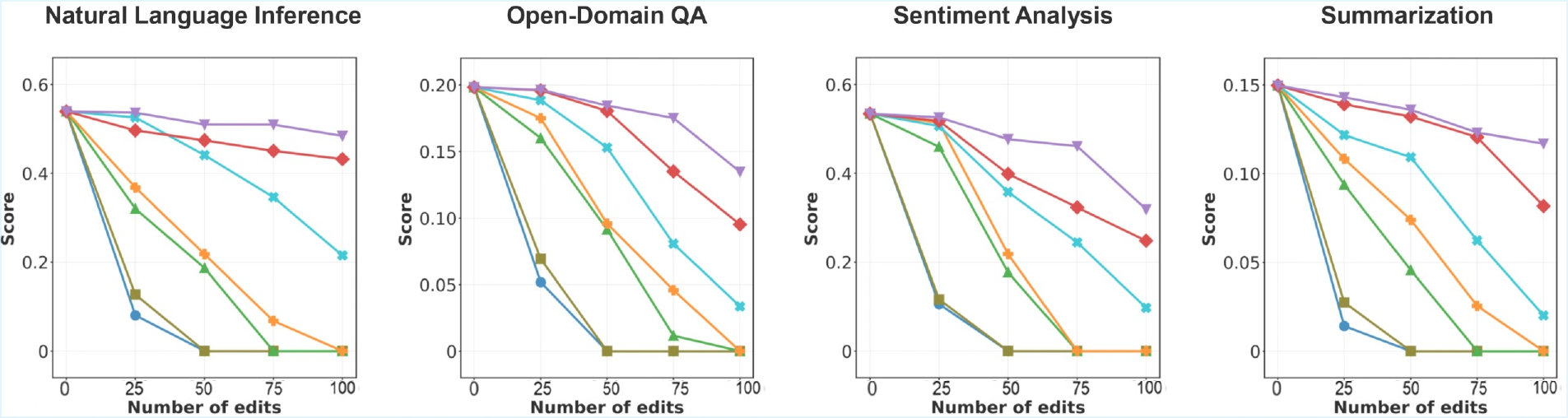}
    \caption{Edited on the ZsRE dataset, the general task performance of varying methods with LLaMA3-8B as the number of edits increases, under the batch-sequential scenario. We set the batch size to 100.}
    \label{batch-sequential-f-zsre-llama3}
\end{figure*}

\begin{figure*}[t]
    \centering
    \includegraphics[width=0.7\linewidth]{figures/legend_new.pdf}
    \includegraphics[width=0.95\linewidth]{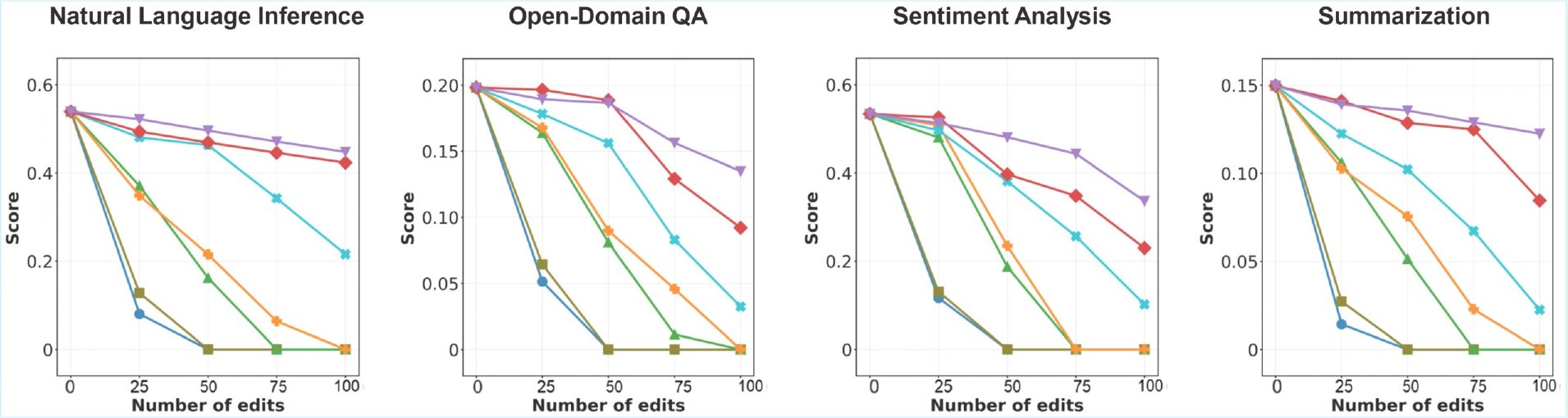}
    \caption{Edited on the ConceptEdit-Intra dataset, the general task performance of varying methods with LLaMA3-8B as the number of edits increases, under the batch-sequential scenario. We set the batch size to 100.}
    \label{batch-sequential-f-intra-llama3}
\end{figure*}

\begin{figure*}[t]
    \centering
    \includegraphics[width=0.9\linewidth]{figures/legend_new.pdf}
    \includegraphics[width=0.95\linewidth]{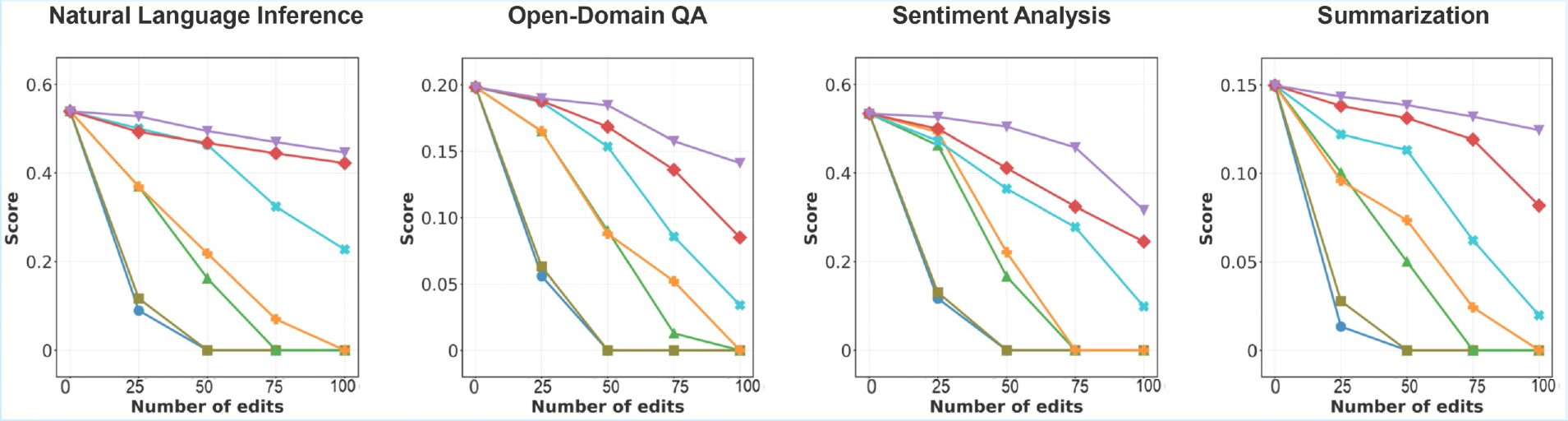}
    \caption{Edited on the ConceptEdit-Inter dataset, the general task performance of varying methods with LLaMA3-8B as the number of edits increases, under the batch-sequential scenario. We set the batch size to 100.}
    \label{batch-sequential-f-inter-llama3}
\end{figure*}

\begin{figure*}[t]
    \centering
    \includegraphics[width=0.9\linewidth]{figures/legend_new.pdf}
    \includegraphics[width=0.95\linewidth]{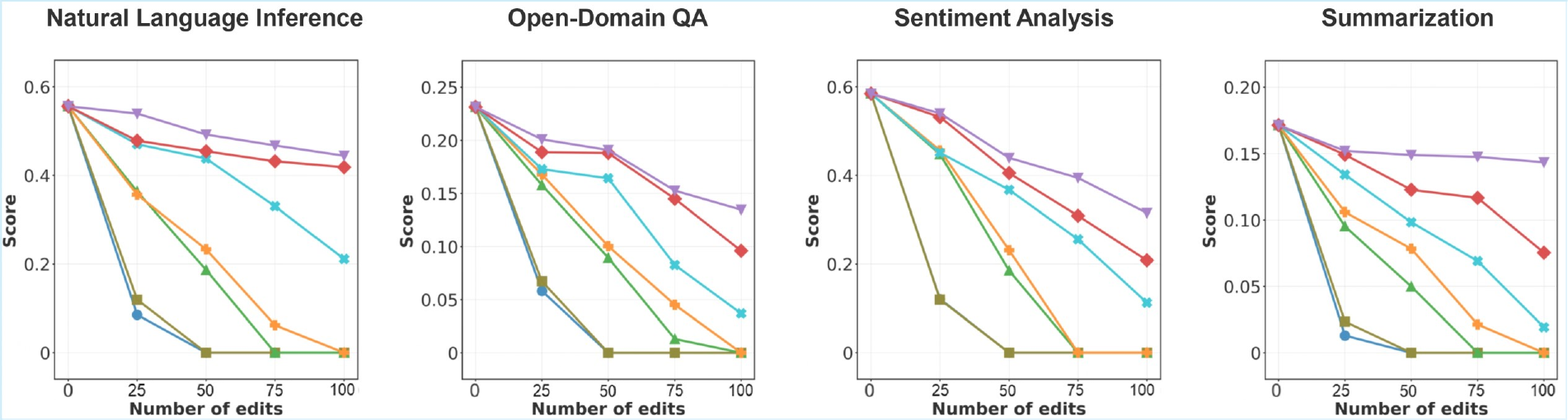}
    \caption{Edited on the ConceptEdit-Inter dataset, the general task performance of varying methods with Qwen2.5-7B as the number of edits increases, under the batch-sequential scenario. We set the batch size to 100.}
    \label{batch-sequential-f-inter-qwen}
\end{figure*}

\begin{figure*}[t]
    \centering
    \includegraphics[width=0.9\linewidth]{figures/legend_new.pdf}
    \includegraphics[width=0.95\linewidth]{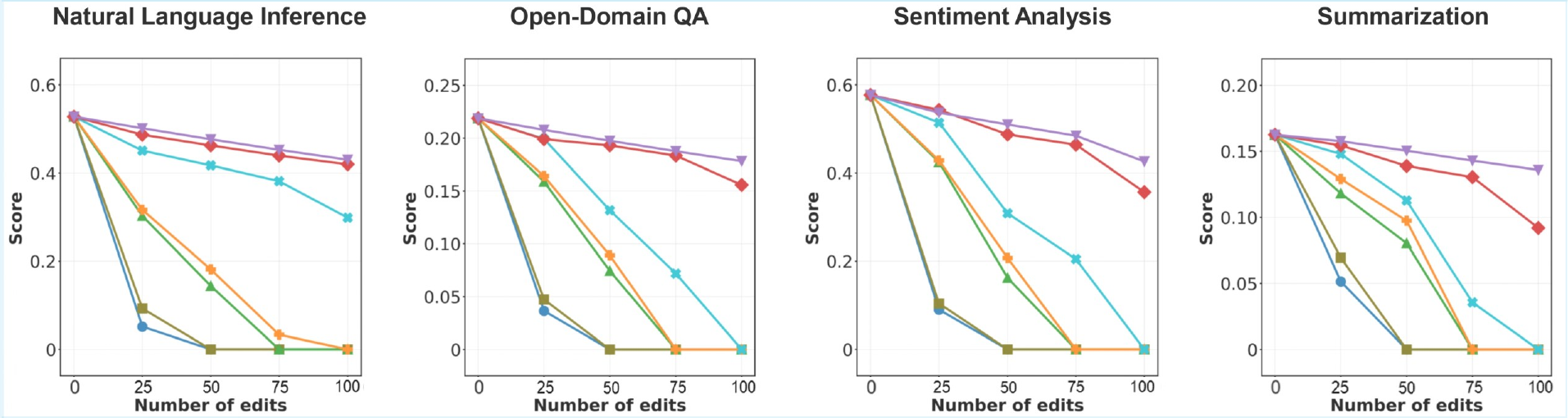}
    \caption{Edited on the ConceptEdit-Inter dataset, the general task performance of varying methods with LLaMA2-13B as the number of edits increases, under the batch-sequential scenario. We set the batch size to 100.}
    \label{batch-sequential-f-inter-llama2}
\end{figure*}

\paragraph{General Abilities}
Figure~\ref{batch-sequential-f-ct-llama2} and Figure~\ref{batch-sequential-f-ct-qwen} illustrate the general abilities of models based on LLaMA2-13B and Qwen2.5-7B on the CounterFact dataset. Figure~\ref{batch-sequential-f-zsre-llama3}, Figure~\ref{batch-sequential-f-intra-llama3} and Figure~\ref{batch-sequential-f-inter-llama3} further show the general abilities of the model using LLaMA3-8B on the ZsRE, ConceptEdit-Intra and ConceptEdit-Inter. 
Similarly, Figure~\ref{batch-sequential-f-inter-qwen} illustrates the general abilities of the Qwen2.5-7B ConceptEdit-Inter datasets, while Figure~\ref{batch-sequential-f-inter-llama2} illustrates the general abilities of the LLaMA2-13B on the same dataset. In our experiments, we set the batch size to 100.
Overall, the results from these experiments are consistent with our earlier observations and provide strong evidence for the effectiveness of \METHODNAME{} in maintaining general abilities.

\section{Analysis of Edited Facts Forgetting and General Ability Decline}\label{fur}
In this section, we extend the analysis by employing visualization techniques to investigate the underlying reasons for the forgetting of edited facts and the degradation of generalization ability after editing. Furthermore, we illustrate how \METHODNAME{} alleviates these challenges and contributes to more stable model performance.
Let the set of editing facts be $\mathcal{E} = \{e_1, e_2, \ldots\}$ with $|\mathcal{E}|=1000$, and the set of downstream prompts be $\mathcal{P} = \{p_1, p_2, \ldots\}$ with $|\mathcal{P}|=1000$, used to evaluate knowledge retention and generalization, respectively. RECT was used for analysis with the initial weight matrix. 

\begin{figure*}[t]
    \centering
    \includegraphics[width=0.6\linewidth]{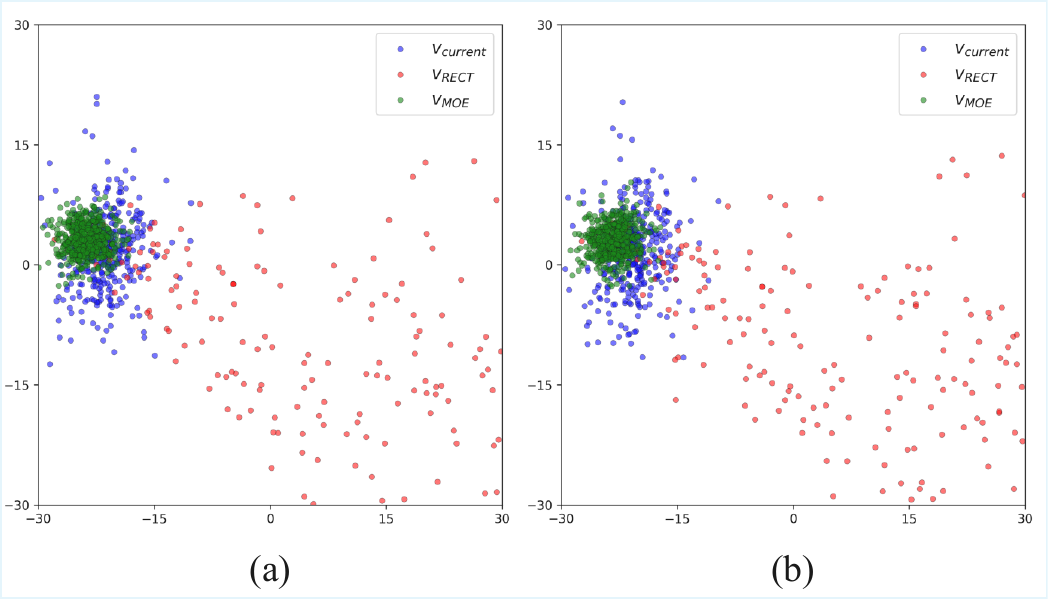}
    \caption{2-dimensional PCA visualization of first 200 values, conducted with LLaMA3-8B on CounterFact.}
    \label{visual}
\end{figure*}

\paragraph{Editing Facts Forgetting}
In sequential editing, RECT derives key–value pairs \((k_j^{e}, v_j^{e})\) for the last subject token and applies updates \(\Delta W_j\) so that \(W_j k_j^{e}=v_j^{e}\).We used the final matrix \(W_{1000}\) and obtained \(W_{1000}k_j^{e}=\hat v_j^{e}\), while with \METHODNAME{} we obtain \(W_{1000}k_j^{e}=\bar v_j^{e}\). If \(\hat v_j^{e}\) stays close to \(v_j^{e}\), the fact \(e_j\) is preserved.
Principal Component Analysis (PCA)~\citep{DBLP:journals/csur/GewersFASCAC21} was employed to visualize these values. The first two principal components of each value were calculated and illustrated, as they can represent most of its features~\citep{DBLP:conf/icml/ZhengY0M0CHP24}.
Let \(V_{\text{current}}=\{v_j^{e}\}\), \(V_{\text{RECT}}=\{\hat v_j^{e}\}\), and \(V_{\text{\METHODNAME{}}}=\{\bar v_j^{e}\}\). As shown in Figure~\ref{visual}(a), PCA shows a large separation between \(V_{\text{current}}\) and \(V_{\text{RECT}}\) but a small gap between \(V_{\text{current}}\) and \(V_{\text{\METHODNAME{}}}\), indicating that \METHODNAME{} mitigates fact forgetting.

\paragraph{General Ability Decline}
We conducted a parallel study for general abilities. Let \(\mathcal{K}=\{k_i^{p}\}\) be task-agnostic keys with baseline outputs \(v_i^{p} = W k_i^{p}\). After sequential editing with RECT, the final matrix gives \(W_{1000}k_i^{p}=\hat v_i^{p}\) and with \METHODNAME{}, \(W_{1000}k_i^{p}=\bar v_i^{p}\). If \(\hat v_i^{p} \approx v_i^{p}\), the corresponding general ability is preserved. Define \(U_{\text{current}}=\{v_i^{p}\}\), \(U_{\text{RECT}}=\{\hat v_i^{p}\}\), and \(U_{\text{\METHODNAME{}}}=\{\bar v_i^{p}\}\). Following the same PCA protocol, as illustrated in Figure~\ref{visual}(b), PCA shows a large separation between \(U_{\text{current}}\) and \(U_{\text{RECT}}\), but only a small gap between \(U_{\text{current}}\) and \(U_{\text{\METHODNAME{}}}\), indicating that RECT induces drift in general representations whereas \METHODNAME{} constrains updates and better preserves general abilities.